\documentclass[nohyperref]{article}

\usepackage{microtype}
\usepackage{graphicx}
\usepackage{subfigure}
\usepackage{booktabs} 

\usepackage{hyperref}

\usepackage[accepted]{icml2022}

\usepackage{amsmath}
\usepackage{amssymb}
\usepackage{mathtools}
\usepackage{amsthm}

\usepackage[utf8]{inputenc}
\usepackage[T1]{fontenc}    
\usepackage{amsfonts}       
\usepackage{nicefrac}       
\usepackage{microtype}      
\usepackage{xcolor}         
\usepackage{natbib}

\usepackage{xspace}

\usepackage{enumerate, paralist}
\usepackage{natbib}
\usepackage{comment}
\usepackage{multirow}
\usepackage{xspace}
\usepackage{rotating}
\usepackage{graphicx}
\usepackage{booktabs} 
\usepackage{tabularx}
\usepackage{enumitem}
\theoremstyle{plain}

\usepackage{hyperref}

\newtheorem{theorem}{Theorem}[section]

\newtheorem{lemma}[theorem]{Lemma}
\newtheorem*{prove*}{Prove}

\newtheorem{assumption}[theorem]{Assumption}

\theoremstyle{definition} 
\newtheorem{definition}[theorem]{Definition}
\newtheorem{remark}[theorem]{Remark}

\renewcommand{\algorithmicrequire}{\textbf{Input:}}
\renewcommand{\algorithmicensure}{\textbf{Output:}}

\newcommand{\bx}{\mathbf{x}}
\newcommand{\BX}{\mathbf{X}}
\newcommand{\bw}{\mathbf{W}}
\newcommand{\bbr}{\mathbb{R}}
\newcommand{\bbe}{\mathbb{E}}

\newcommand{\caln}{\mathcal{N}}
\newcommand{\para}[1]{\noindent \textbf{#1}\xspace}

\newcommand{\cald}{\mathcal{D}}
\newcommand{\calo}{\mathcal{O}}
\newcommand{\calt}{\mathcal{T}}
\newcommand{\call}{\mathcal{L}}

\newcommand{\sysn}{\text{Meta-Ban}\xspace}

\title{Neural Collaborative Filtering Bandits via Meta Learning}
\date{}

\author{Yikun Ban, Yunzhe Qi, Tianxin Wei, Jingrui He \\
University of Illinois Urbana-Champaign    }

\begin{document}
\maketitle

\begin{abstract}

Contextual multi-armed bandits provide powerful tools to solve the exploitation-exploration dilemma in decision making, with direct applications in the personalized recommendation. In fact, collaborative effects among users carry the significant potential to improve the recommendation. In this paper, we introduce and study the problem by exploring `Neural Collaborative Filtering Bandits', where the rewards can be non-linear functions
and groups are formed dynamically given different specific contents. To solve this problem, inspired by meta-learning, we propose Meta-Ban (\textbf{meta-ban}dits), where a meta-learner is designed to represent and rapidly adapt to dynamic groups, along with a UCB-based exploration strategy. Furthermore, we analyze that Meta-Ban can achieve the regret bound of $\mathcal{O}(\sqrt{T \log T})$, improving a multiplicative factor $\sqrt{\log T}$ over state-of-the-art related works. In the end, we conduct extensive experiments showing that Meta-Ban significantly outperforms six strong baselines.
\end{abstract}

\vspace{-0.3cm}
\section{Introduction}

The contextual multi-armed bandit has been extensively studied in machine learning to resolve the exploitation-exploration dilemma in sequential decision making, with wide applications in personalized recommendation \citep{2010contextual}, online advertising~\cite{wu2016contextual}, etc.

Recommender systems play an indispensable role in many online businesses, such as e-commerce providers and online streaming services. It is well-known that the collaborative effects are strongly associated with user preference. Thus, discovering and leveraging collaborative information in recommender systems has been studied for decades. In the relatively static environment, e.g., in a movie recommendation platform where catalogs are known and accumulated ratings for items are provided, the classic collaborative filtering can be easily deployed (e.g., matrix/tensor factorization \citep{su2009survey}). However, such methods can hardly adapt to more dynamic settings, such as news or short-video recommendation, due to: (1) the lack of cumulative interactions for new users or items; (2) the difficulty of balancing the exploitation of current user-item preference knowledge and exploration of the new potential matches (e.g., presenting new items to the users).

To address this problem, a line of works, clustering of bandits (collaborative filtering bandits) \citep{2014onlinecluster,2016collaborative,gentile2017context,2019improved,ban2021local}, have been proposed to incorporate collaborative effects among users which are largely neglected by conventional bandit algorithms \citep{dani2008stochastic,2011improved,valko2013finite,ban2020generic}. These works adaptively cluster users and explicitly or implicitly utilize the collaborative effects on both user and arm (item) sides while selecting an arm. However, this line of works have a significant limitation that they all build on the linear bandit framework \citep{2011improved}. This linear reward assumption may not be true in real-world applications \citep{valko2013finite}.

To learn non-linear reward functions, neural bandits \citep{zhou2020neural,zhang2020neural} have attracted much attention, where a neural network is assigned to learn the reward function along with an exploration strategy (e.g., Upper Confidence Bound (UCB) and Thompson Sampling (TS)). However, this class of works do not incorporate any collaborative effects among users, overlooking the crucial potential in improving recommendation.

In this paper, to overcome the above challenges, we first introduce the problem, Neural Collaborative Filtering Bandits (NCFB), built on either linear or non-linear reward assumptions while introducing relative groups. Groups are formed by users sharing similar interests/preferences/behavior. However, such groups usually are not static over specific contents. For example, two users may both like "country music" but may have different opinions on "rock music". "Relative groups" are introduced in NCFB to formulate groups given a specific content, which is more practical in real problems.

To solve NCFB, the bandit algorithm has to show representation power to formulate user/group behavior and strong adaptation in matching rapidly-changing groups. Therefore, we propose a bandit algorithm, \sysn (\textbf{Meta-Ban}dits), inspired by recent advances in meta-learning \cite{finn2017model, yao2019hierarchically}. In \sysn, a meta-learner is assigned to represent and rapidly adapt to dynamic groups. And a user-learner is assigned to each user to discover the underlying relative groups. We use a neural network to model both meta-learner and user learners, in order to learn linear or non-linear reward functions. To solve the exploitation-exploration dilemma in bandits, \sysn has a UCB-based strategy for exploration,
In the end, we provide rigorous regret analysis and empirical evaluation for \sysn. 
This is the first work incorporating collaborative effects in neural bandits to the best of our knowledge.  
The contributions of this paper can be summarized as follows:

\begin{enumerate}
\vspace{-0.4cm}
    \item \textbf{Problem.} We introduce the problem, Neural Collaborative Filtering Bandits (NCFB), to incorporate collaborative effects among users with either linear or non-linear reward assumptions.
    \vspace{-0.3cm}
    \item \textbf{Algorithm.} We propose a neural bandit algorithm working in NCFB, \sysn, where the meta-learner is introduced to represent and rapidly adapt to dynamic groups, along with a new informative UCB for exploration.
    \vspace{-0.3cm}
    \item \textbf{Theoretical analysis.} Under the standard assumption of over-parameterized neural networks, we prove that $\sysn$ can achieve the regret upper bound, $\mathcal{O}(\sqrt{T \log T})$, improving by a multiplicative factor of $\sqrt{\log T}$ over existing state-of-the-art bandit algorithms.
    This is the first near-optimal regret bound in neural bandits incorporating meta-learning to the best of our knowledge. 
    Furthermore, we provide the convergence and generalization bound of meta-learning in the bandit framework, and the correctness guarantee of captured groups, which may be of independent interests. 
    \vspace{-0.3cm}
    \item \textbf{Empirical performance.} We evaluate \sysn on four real-world datasets and show that \sysn outperforms six strong baselines.
\end{enumerate}
\vspace{-0.4cm}
Next, after briefly reviewing related works in Section \ref{sec:related}, we show the problem definition in Section \ref{sec:prod} and introduce the proposed \sysn in Section \ref{sec:algorithm} together with theoretical analysis in Section \ref{sec:theo1}-\ref{sec:detail}. In the end, we present the experiments in Section \ref{sec:exp} and conclusion in Section \ref{sec:conclusion}.

\vspace{-0.3cm}

\section{Related Work} \label{sec:related}

In this section, we briefly review the related works, including clustering of bandits and neural bandits.

\para{Clustering of bandits.} CLUB~\cite{2014onlinecluster} first studies exploring collaborative knowledge among users in contextual bandits where each user hosts an unknown vector to represent the behavior based on the linear reward function. CLUB formulates user similarity on an evolving graph and selects an arm leveraging the clustered groups. Then, \citet{2016collaborative,gentile2017context} propose to cluster users based on specific contents and select arms leveraging the aggregated information of conditioned groups. \citet{2019improved} improve the clustering procedure by allowing groups to split and merge. \citet{ban2021local} use seed-based local clustering to find overlapping groups, different from globally clustering on graphs. \citet{korda2016distributed,yang2020exploring,wu2021clustering} also study clustering of bandits with various settings in recommendation system. However, all the series of works are based on the linear reward assumption, which may fail in many real-world applications. 

\para{Neural bandits.} \citet{allesiardo2014neural} use a neural network to learn each action and then selects an arm by the committee of networks with $\epsilon$-greedy strategy. \citet{lipton2018bbq, riquelme2018deep} adapt the Thompson Sampling to the last layer of deep neural networks to select an action. However, these approaches do not provide regret analysis. \citet{zhou2020neural,ban2021convolutional} and  \citet{zhang2020neural} first provide the regret analysis of UCB-based and TS-based neural bandits, where they apply ridge regression on the space of gradients. \citet{ban2021multi} study a combinatorial problem in multiple neural bandits with a UCB-based exploration.
EE-Net\citep{ban2021ee} proposes to use another neural network for exploration.
Unfortunately, all these methods neglect the collaborative effects among users in contextual bandits.
\vspace{-0.3cm}
\section{Neural Collaborative Filtering Bandits} \label{sec:prod}

In this section, we introduce the problem of Neural Collaborative Filtering bandits, motivated by generic recommendation scenarios. 

Suppose there are  $n$ users, $N = \{1, \dots, n\}$, to serve on a platform. In the $t^{\textrm{th}}$ round, the platform receives a user $u_t \in N$ and prepares the corresponding $k$ arms (items) $\BX_t = \{ \bx_{t, 1}, \bx_{t, 2}, \dots, \bx_{t,k} \}$ in which each arm is represented by its $d$-dimensional feature vector  $\bx_{t, i} \in \bbr^d, \forall i \in \{1, \dots, k\}$. Then, like the conventional bandit problem, the platform will select an arm $\bx_{t,i} \in \BX_t$ and recommend it to the user $u_t$. In response to this action, $u_t$ will produce a corresponding reward (feedback) $r_{t,i}$.   We use $r_{t,i}| u_t$ to represent the reward produced by $u_t$ given $\bx_{t,i}$, because different users may generate different rewards towards the same arm.

Group behavior (collaborative effects) exists among users and has been exploited in recommender systems. In fact, the group behavior is item-varying, i.e., the users who have the same preference on a certain item may have different opinions on another item. Therefore, we define a {\it relative group} as a set of users with the same opinions on a certain item.
\vspace{-0.3cm}
\begin{definition}[Relative Group]  \label{def:group}
In round $t$, given an arm $\bx_{t,i} \in \BX_t$, a relative group $\caln(\bx_{t,i}) \subseteq N$ with respect to $\bx_{t,i}$ satisfies
\[
\begin{aligned}
    &1) \ \forall u, u' \in \caln(\bx_{t,i}),   \bbe[r_{t,i}| u] = \bbe[r_{t,i}| u']\\
    &2) \  \nexists \ \caln \subseteq N,  \text{s.t.}   \ \caln \ \text{satisfies} \ 1) \ \text{and} \  \caln(\bx_{t,i}) \subset \caln.
\end{aligned}
\]
\end{definition}
Such flexible group definition allows users to agree on certain items while disagree on others, which is consistent with the real-world scenario.

Therefore, given an arm $\bx_{t,i}$, the user pool $N$ can be divided into $q_{t,i}$ non-overlapping  groups: $\caln_1(\bx_{t,i}), \caln_2(\bx_{t,i}),$ $\dots, \caln_{q_{t,i}}(\bx_{t,i})$, where $q_{t,i} < n$.
Note that the group information is \textbf{unknown} to the platform.
We expect that the users from different groups have distinct behavior with respect to $\bx_{t,i}$. Thus, we provide the following constraint among groups.

\begin{definition} [$\gamma$-gap]  \label{def:gap}
Given two different groups $\caln(\bx_{t,i})$, $\caln'(\bx_{t,i})$, they satisfy
\[
\forall u \in \caln(\bx_{t,i}), u' \in \caln'(\bx_{t,i}),   |  \bbe[r_{t,i}| u] -   \bbe[r_{t,i}| u'] | \geq \gamma.
\]
\end{definition}

For any two groups in $N$, we assume that they satisfy the $\gamma$-gap constraint. Note that such an assumption is standard in the literature of online clustering of bandit to differentiate groups \cite{2014onlinecluster, 2016collaborative, gentile2017context, 2019improved, ban2021local}.

\para{Reward function}. The reward $r_{t,i}$ is assumed to be governed by a universal function with respect to $\bx_{t,i}$ given $u_t$:
 \begin{equation}
 r_{t,i}|u_t =   h_{u_t}(\bx_{t,i}) + \zeta_{t, i},
 \end{equation}
where  $h_{u_t}$ is an either linear or non-linear but unknown reward function associated with $u_t$, and $\zeta_{t, i}$ is a noise term with zero expectation $\bbe[\zeta_{t, i}] = 0$. We assume the reward $r_{t,i}$ is bounded, $r_{t,i} \in [0, 1]$, as many existing works \citep{2014onlinecluster,gentile2017context,ban2021local}. Note that online clustering of bandits  assume $h_{u_t}$ is a linear function with respect to $\bx_{t,i}$ \citep{2014onlinecluster, 2016collaborative, gentile2017context, 2019improved, ban2021local}. 

\para{Regret analysis}. In this problem, the goal is to minimize the expected accumulated regret of $T$ rounds:
\begin{equation} \label{eq:regret}
R_T = \sum_{t=1}^T \bbe[ r_t^\ast  - r_t |u_t, \BX_t],  
\end{equation}
where $r_t$ is the reward received in round $t$ and $\bbe[r_t^\ast |u_t, \BX_t] = \max_{\bx_{t, i} \in \BX_t} h_{u_t}(\bx_{t,i})$.  

The above introduced framework can naturally formulate many recommendation scenarios. For example, for a music streaming service provider, when recommending a song to a user, the platform can exploit the knowledge of other users who have the same opinions on this song, i.e., all `like' or `dislike' this song. Unfortunately, the potential group information is usually not available to the platform before the user's feedback. In the next section, we will introduce an approach that can infer and exploit such group information to improve the recommendation.

\para{Notation.} Denote by $[k]$ the sequential list $\{1, \dots,  k\}$.  
Let $\bx_t$ be the arm selected in round $t$ and $r_t$ be the reward received in round $t$.
We use $\|\bx_t\|_2$ and $\|\bx_t\|_1$ to represent the Euclidean norm and Taxicab norm.
For each user $u \in N$, let $\mu_t^u$ be the number of rounds in which $u$ has occurred up to round $t$, i.e., $\mu_t^u = \sum_{\tau =1}^t \mathbb{I}[u_\tau = u]$, and $\calt^u_{t}$ be all of $u$'s historical data up to round $t$, denoted by $\calt^u_{t} = \{\bx_\tau^u, r_\tau^u \}_{\tau = 1}^{\mu_t^u}$ where $\bx_\tau^u$ is the $\tau$-th arm selected by $u$.
Given a group $\caln$, all it's data up to $t$ can be denote by $ \{\calt^u_t \}_{u \in N}  =  \{ \calt_t^u | u \in \caln \}$. We use standard $\mathcal{O}, \boldsymbol{\Theta}, $ and $\Omega$ to hide constants.
\section{Proposed Algorithm} \label{sec:algorithm}

In this section, we propose a meta-learning-based bandit algorithm, \sysn, to tackle the challenges in the NCFB problem as follows:
\vspace{-0.4cm}
\begin{itemize}
    \item Challenge 1 (C1): Given an arm,  how to infer a user's relative group, and whether the returned group is the true relative group?   
    \item Challenge 2 (C2): Given a relative group, how to represent the group's behavior in a parametric way? 
    \item Challenge 3 (C3): How to generate a model to efficiently adapt to the rapidly-changing relative groups? 
    \item Challenge 4 (C4): How to balance between exploitation and exploration in bandits with relative groups?
        \vspace{-0.3cm}
\end{itemize}
\vspace{-0.3cm}
To represent group/user behavior, \sysn has one meta-learner $\Theta$ and  $n$ user-learners for each user respectively, $\{\theta^u \}_{u \in N}$, sharing the same neural network $f$.
We divide the presentation of \sysn into three parts as follows.

\begin{algorithm}[t]
\caption{ \sysn }\label{alg:main}
\begin{algorithmic}[1]
\STATE {\bfseries Input:}  $T, \nu(\text{group parameter}), \gamma, \alpha (\text{exploration parameter})$,  $\lambda$,  $\delta, J_1, J_2, \eta_1 (\text{user step size}), \eta_2 (\text{meta step size})$.

\STATE Initialize $\Theta_0$;     $\theta_0^u, \widehat{\theta}_0^u =\Theta_0, \mu_0^{u} = 0, \forall u \in N$
\STATE Observe one data for each $ u \in N$
\FOR{ each $t = 1, 2, \dots, T$}
\STATE Receive a user $u_t \in N$ and observe $k$ arms $\BX_t =\{\bx_{t, 1}, \dots, \bx_{t, k}\}$
\FOR{ each $i \in [k]$ }
\STATE Determine $u_t $'s relative groups: $\widehat{\caln}_{u_t}(\bx_{t,i}) =$
\[
\hspace*{-1cm}  \{ u \in N \ | \ |f(\bx_{t,i};  \theta^u_{t-1}) -  f(\bx_{t,i}; \theta^{u_t}_{t-1})| \leq \frac{\nu-1}{\nu} \gamma \}
\]
\STATE  $ \Theta_{t,i}  =  \text{GradientDecent\_Meta} \left(\widehat{\caln}_{u_t}(\bx_{t,i}) \right)   $ 
\STATE  $\text{U}_{t, i} = f(\bx_{t,i}; \Theta_{t,i})  +  \alpha \cdot \text{UCB}_{t,i} $ (Lemma \ref{flemma:ucb}) \ where
\[
\hspace*{-1cm} \text{UCB}_{t,i} = \beta_2 \cdot  \| g(\bx_{t,i}; \Theta_{t, i}) - g(\bx_{t,i}; \theta_{0}^{u_t})\|_2 + Z_1 + \bar{U}_{u_t} \ 
\]
\ENDFOR
\STATE $ i' = \arg_{ i \in [k]} \max  \text{U}_{t, i}    $
\STATE Play $\bx_{t, i'}$ and observe reward $r_{t, i'}$
\STATE $\bx_t =   \bx_{t, i'} ;  \ r_t  =  r_{t, i'}; \   \Theta_t = \Theta_{t, i'}$
\STATE $\mu_t^{u_t} = \mu_{t-1}^{u_t} + 1$
\STATE $\theta^{u_t}_t = \text{GradientDecent\_User}(u_t)$
\FOR{ $u \in N$ and $u \not = u_t$}
\STATE  $\theta^{u}_t =\theta^{u}_{t-1}$; \ $\mu_t^{u} = \mu_{t-1}^{u}$
\ENDFOR
\ENDFOR
\end{algorithmic}
\end{algorithm}

\para{Group inference (to C1).}  
As defined in Section \ref{sec:prod}, each user $u \in N$ is governed by a universal unknown function $h_u$. It is natural to use the universal approximator \citep{hornik1989multilayer}, a neural network $f$, to learn $h_u$. 
In round $t \in [T]$, let $u_t$ be the user to serve. Given $u_t$'s past data up to round $t-1$, $\calt_{t-1}^{u_t}$, we can  train parameters $\theta^{u_t}$  by minimizing the following loss: 
\begin{equation}
\call \left(\calt_{t-1}^{u_t}; \theta^{u_t}  \right) =  \frac{1}{2} \sum_{(\bx, r) \in  \calt_{t-1}^{u_t} }  ( f(\bx; \theta^{u_t}) - r)^2. 
\end{equation}
Let $\theta_{t-1}^{u_t}$ represent $\theta_{u_t}$ trained with $\calt_{t-1}^{u_t}$ in round $t-1$.
The training of $\theta^{u_t}$ can be conducted by (stochastic) gradient descent, e.g., as described in Algorithm \ref{alg:user},  where $\theta^{u}_t$ is uniformly drawn from  $u$'s historical parameters $\{\widehat{\theta}_\tau^u\}_{\tau = 0}^{\mu^t_u}$ to obtain a theoretical generalization bound (refer to Lemma \ref{flemma:usergeneral}).

Therefore, for each $u \in N$, we can obtain the trained parameters $\theta^u_{t-1}$. Then,
given $u_t$ and an arm $\bx_{t,i}$, we return $u_t$'s estimated group with respect to an arm $\bx_{t,i}$ by 
\begin{equation}
\begin{aligned}
&\widehat{\caln}_{u_t}(\bx_{t,i}) = \\
\{ u \in N \ | \ |f(\bx_{t,i}; & \theta^u_{t-1}) -  f(\bx_{t,i}; \theta^{u_t}_{t-1})| \leq \frac{\nu-1}{\nu} \gamma \}.
\end{aligned}
\end{equation}
where $\gamma \in [0, 1)$ represents the assumed $\gamma$-gap and $\nu > 1$ is a tuning parameter to trade off between the exploration of group members and the cost of playing rounds. 

We emphasize that $\caln_{u_t}(\bx_{t,i})$ ($u_t \in \caln_{u_t}(\bx_{t,i})$) is the ground-truth relative group satisfying Definition \ref{def:group}. Suppose $\gamma$-gap holds among $N$, we prove that when $t$ is larger than a constant, i.e., $t \geq \widetilde{T}$, with probability at least $1  - \delta$,  it holds uniformly that $ \widehat{\caln}_{u_t}(\bx_{t,i}) = \caln_{u_t}(\bx_{t,i})$ (Lemma \ref{flemma:groupgurantee}). 
Then, for $\nu$, we have: 
(1) When $\nu \uparrow$,  we have more chances to explore collaboration with other users while costing more rounds ($\widetilde{T} \uparrow$); (2) When $\nu \downarrow$,  we limit the potential cooperation with other users while saving exploration rounds ($\widetilde{T} \downarrow$).

\begin{algorithm}[t]
\caption{GradientDecent\_Meta ($\caln$) }\label{alg:meta}
\begin{algorithmic}[1]
\STATE $\Theta_{(0)} =  \Theta_{0}$ (or  $\Theta_{t-1}$) \ \
\FOR{$ j = 1, 2, \dots, J_2$}
\FOR{ $u \in \caln$}
\STATE Collect $ \mathcal{T}^{u}_{t-1}$ 
\STATE Randomly choose $\widetilde{\calt}^u  \subseteq \mathcal{T}^{u}_{t-1}$
\STATE $\call \left(\widehat{\theta}^u_{\mu^u_{t-1}} \right) =  \frac{1}{2} \sum_{(\bx, r) \in  \widetilde{\calt}^u  }  ( f(\bx; \widehat{\theta}^u_{\mu^u_{t-1}}) - r)^2  $
\ENDFOR
\STATE $\call_{\caln}=  \sum_{u \in \caln} w_{u} \cdot \call \left( \widehat{\theta}^u_{\mu^u_{t-1}}  \right) + \frac{\lambda}{\sqrt{m}} \sum_{u \in \caln}  \| \widehat{\theta}^u_{\mu^u_{t-1}} \|_1. $    
\STATE $\Theta_{(j)} = \Theta_{(j-1)} -  \eta_2 \triangledown_{ \{ \widehat{\theta}^u_{\mu^u_{t-1}} \}_{u \in \caln}  }\call_\caln$
\ENDFOR
\STATE \textbf{Return:} $\Theta_{(J_2)}$
\end{algorithmic}
\end{algorithm}

\para{Meta learning (to C2 and C3)}. In this paper, we propose to use one meta-learner $\Theta$ to represent and adapt to the behavior of dynamic groups.
In meta-learning, the meta-learner is trained based on a number of different tasks and can quickly learn new tasks from small amount of new data \citep{finn2017model}.
Here, we consider each user $u \in N$ as a task and its collected data $\calt_t^u$ as the task distribution. 
Therefore, \sysn has two phases: User adaptation and Meta adaptation.

\textit{User adaptation}. In the $t^{\textrm{th}}$ round, given $u_t$,  after receiving the reward $r_t$, we have available data $\calt_t^{u_t}$. Then, the user parameter $\theta^{u_t}$ is updated in round $t$ based on meta-learner $\Theta$, denoted by $\theta^{u_t}_{t}$, described in Algorithm \ref{alg:user}.

\textit{Meta adaptation}. In the $t^{\textrm{th}}$ round, given a group $\widehat{\caln}_{u_t}(\bx_{t,i})$,
we have the available collected data $\{ \calt_{t-1}^u \}_{u \in \widehat{\caln}_{u_t}(\bx_{t,i})}$.
The goal of meta-learner is to fast adapt to these users (tasks). Thus, given an arm $\bx_{t,i}$,  we update $\Theta$ in round $t$, denoted by $\Theta_{t, i}$, by minimizing the following meta loss:
\begin{equation}
\begin{aligned}
\call_{\widehat{\caln}_{u_t}(\bx_{t,i})}  = & \sum_{u \in \widehat{\caln}_{u_t}(\bx_{t,i})} w_{u} \cdot \call \left(\widehat{\theta}^u_{\mu^u_{t-1}}  \right) \\
&  +  \frac{\lambda}{\sqrt{m}} \sum_{u \in \widehat{\caln}_{u_t}(\bx_{t,i})} \| \widehat{\theta}^u_{\mu^u_{t-1}} \|_1.  
\end{aligned}
\end{equation}
where $ \widehat{\theta}^u_{\mu^u_{t-1}}$ are the stored user parameters in Algorithm \ref{alg:user} at round $t-1$,  the second term is the L1-Regularization, and $w_u\geq 0$ is the weight for $u$ to adjust its contribution in this update. Usually, we want $w_{u_t}$ to be larger than others. Then, the meta learner is updated by:
\begin{equation}
\Theta = \Theta -  \eta_2 \triangledown_{\{ \widehat{\theta}^u_{\mu^u_{t-1}}\}_{u \in \widehat{\caln}_{u_t}(\bx_{t,i})}} \call_{\widehat{\caln}_{u_t}(\bx_{t,i})}. 
\end{equation}
where $\eta_2$ is the meta learning rate.
Algorithm \ref{alg:meta} shows meta update with stochastic gradient descent (SGD).


\begin{algorithm}[t]
\caption{GradientDecent\_User ($u$) }\label{alg:user}
\begin{algorithmic}[1]
\STATE Collect $\calt^u_{t}$  \ \   \# Historical data of $u$ up to round $t$ 
\STATE $\theta^u_{(0)} = \Theta_0 $ ( or $\Theta_t$)  
\FOR{ $j = 1, 2, \dots, J_1$}
\STATE Randomly choose $\widetilde{\calt}^u  \subseteq \mathcal{T}^{u}_t$
\STATE $\call \left(\widetilde{\calt}^u; \theta^u  \right) =  \frac{1}{2} \sum_{(\bx, r) \in  \widetilde{\calt}^u  }  ( f(\bx; \theta^u) - r)^2  $
\STATE $\theta^u_{(j)} = \theta_{(j-1)}^u -  \eta_1 \triangledown_{\theta^u_{(j-1)}} \call$
\ENDFOR
\STATE $\widehat{\theta}^u_{\mu^u_t} = \theta^u_{(J_1)}$
\STATE \textbf{Return:} Choose $\theta^u_t$ uniformly from $\{\widehat{\theta}_0^u, \widehat{\theta}^u_1, \dots, \widehat{\theta}^u_{\mu^u_t}\}$ 
\end{algorithmic}
\end{algorithm}

\para{UCB Exploration (to C4)}.
To trade off between the exploitation of the current group information and the exploration of new matches, we introduce the following UCB-based selection criterion.

Given the user $u_t$, an arm $\bx_{t,i}$, and the estimated group $\widehat{\caln}_{u_t}(\bx_{t,i})$, based on Lemma \ref{flemma:ucb}, with probability at least $1-\delta$, it holds that:
\begin{equation}
|h_{u_t}(\bx_{t,i}) - f(\bx_{t,i}; \Theta_{t,i}) | \leq  \text{UCB}_{t,i}
\end{equation}
where $\text{UCB}_{t,i} = \beta_2 \cdot  \| g(\bx_{t,i}; \Theta_{t, i}) - g(\bx_{t,i}; \theta_{0}^{u_t})\|_2 + Z_1 + \bar{U}_{u_t},$ defined in Section \ref{sec:detail}.
Note that this UCB contains both meta-side ($g(\bx_{t,i}; \Theta_{t, i})$) and user-side ($\bar{U}_{u_t}$) information to help \sysn leverage the collaborative effects existed in $\widehat{\caln}_{u_t}(\bx_{t,i})$ and $u_t$'s personal behavior to make explorations.

Then, we select an arm according to:
\begin{equation}  \label{eq:selectioncri}
\bx_t = \arg_{ \bx_{t,i} \in \BX_t } \max f(\bx_{t,i}; \Theta_{t,i})  + \alpha \cdot \text{UCB}_{t,i}. 
\end{equation}

To sum up, Algorithm \ref{alg:main} depicts the workflow of \sysn. In each round, given a served user and a set of arms, we compute the meta-learner and its UCB for the relative group with respect to each arm. Then, we choose the arm according to Eq.(\ref{eq:selectioncri}) (Lines 4-12). After receiving the reward, we update the user-learner $\theta^{u_t}$ (Line 14-15) because only $u_t$'s collected data is updated. In the end, we update all the other parameters (Lines 16-18).

\section{Theoretical Analysis} \label{sec:theo1}

In this section, we provide the regret analysis of \sysn and the comparison with existing works.

The analysis focuses on the over-parameterized neural networks \citep{ntk2018neural, allen2019convergence} as other neural bandits \citep{zhou2020neural, zhang2020neural}. 

Given an arm $\bx \in \bbr^d$, without loss of generality, we define $f$ as a fully-connected network with depth $L \geq 2$ and width $m$:
\begin{equation} \label{eq:structure}
f(\bx; \theta \ \text{or} \ \Theta  ) = \bw_L \sigma ( \bw_{L-1}  \sigma (\bw_{L-2} \dots  \sigma(\bw_1 \bx) ))
\end{equation}
where $\sigma$ is the ReLU activation function,  $\bw_1 \in \bbr^{m \times d}$, $ \bw_l \in \bbr^{m \times m}$, for $2 \leq l \leq L-1$, $\bw^L \in \bbr^{1 \times m}$, and  $\theta, \Theta  = [ \text{vec}(\bw_1)^\intercal,  \text{vec}(\bw_2)^\intercal, \dots, \text{vec}(\bw_L )^\intercal ]^\intercal \in \bbr^{p}$. To conduct the analysis, we need the following initialization and  mild assumptions.

\textit{Initialization}.   For $l \in [L-1]$, each entry of $\bw_l$ is drawn from the normal distribution $\caln(0, 2/m)$; Each entry of $\bw_L$ is drawn from the normal distribution $\caln(0, 1/m)$.

\begin{assumption} [Distribution]  \label{assum:1}
All arms and rewards are assumed to be drawn, i.i.d, from the general distribution, $\cald$. Specifically, in each round $t$, the serving user $u_t$ and the data distribution with respect to $u_t$ are drawn from $\cald$, i.e., $(u_t, \cald_{u_t} ) \sim \cald$. Then, in the $t^{\textrm{th}}$ round, for each $i \in [k]$,  the arm and reward $(\bx_{t,i}, r_{t, i})$ with respect to $u_t$ are drawn from the data distribution $\cald_{u_t}$, i.e., $(\bx_{t,i}, r_{t, i}) \sim \cald_{u_t}, \forall i \in [k]$.
\end{assumption}

\begin{assumption} [Arm Separability]  \label{assum:2}
For $t \in [T], i \in[k], \|\bx_{t, i}\|_2 = 1$. 
Then, for every pair $\bx_{t, i}, \bx_{t', i'}$, $t' \in[T], i' \in [k],$  and $(t, i) \neq (t', i')$,  $\|\bx_{t, i} - \bx_{t', i'}\|_2 > \rho$.
\end{assumption}

Assumption \ref{assum:1} is the standard task distribution in meta learning \citep{wang2020onglobal,wang2020globalcon} and Assumption \ref{assum:2} is the standard input assumption in over-parameterized neural networks \cite{allen2019convergence}. Then, we provide the following regret upper bound  for \sysn with gradient descent. 

\begin{theorem} \label{theorem:main}
Given the number of rounds $T$, assume that $(u_t, \cald_{u_t})$ is uniformly drawn from $\cald$. For any $\delta \in (0, 1), \lambda > 0,  0 < \epsilon_1 \leq \epsilon_2 \leq 1, \rho \in (0, \calo(\frac{1}{L}))$, suppose $m, \eta_1, \eta_2, J_1, J_2$ satisfy
\begin{equation} \label{eq:conditions0}
\begin{aligned}
m &\geq \widetilde{\Omega} \left( \max \left\{  \text{poly}(t, L, \rho^{-1}), e^{\sqrt{\log ( \calo(Tk)/ \delta)}} \right \} \right) \\ 
\eta_1 & = \boldsymbol{\Theta} \left(  \frac{\rho}{ \text{poly}(t, L) m}\right), \ \
J_1 = \boldsymbol{\Theta} \left(  \frac{\text{poly}(t, L)}{\rho^2} \log \frac{1}{\epsilon_1}\right) \\
\eta_2& =  \min  \left \{  \boldsymbol{\Theta}\left( \frac{\sqrt{n}\rho}{ t^4 L^2 m} \right),    \boldsymbol{\Theta} \left( \frac{\sqrt{\rho \epsilon_2}}{ t^2 L^2 \lambda} n^2 \right) \right \} ,  \\
 J_2& = \max \Bigg \{  \boldsymbol{\Theta}\left(  \frac{t^5(\calo(t \log^2 m) - \epsilon_2) L^2 m}{ \sqrt{n \epsilon_2}\rho }\right),  \\
& \boldsymbol{\Theta}\left( \frac{t^3 L^2 \lambda n^2 (\calo(t \log^2 m - \epsilon_2)) }{\rho \epsilon_2}   \right)  \Bigg \}.
\end{aligned}
\end{equation}
Set $w_u = 1, \widetilde{\calt}^u  = \mathcal{T}^{u}_t$ for $u \in N, t \in [T]$.
Then, with probability at least $1 - \delta$ over the initialization of $\Theta_0$, Algorithms \ref{alg:main}-\ref{alg:user} have the following regret upper bound:
\[
\begin{aligned}
R_T \leq &   \calo \left( (2 \sqrt{T} -1) \sqrt{n}  + \sqrt{2 T \log (1/\delta)} \right)  \\
& \cdot \calo( \sqrt{ \log \frac{\calo(Tk)}{ \delta}} ).
\end{aligned}
\]
\end{theorem}

\para{Comparison with clustering of bandits}. The existing works on clustering of bandits \citep{2014onlinecluster,2016collaborative,gentile2017context,2019improved,ban2021local} are all based on the linear reward assumption and achieve the following regret bound complexity:
\[
R_T  \leq  \calo( d \sqrt{T} \log T).
\]

\para{Comparison with neural bandits}. The regret analysis in a single neural bandit  \citep{zhou2020neural,zhang2020neural} has been developed recently, achieving
\[
R_T  \leq  \calo(\tilde{d} \sqrt{T} \log T) \ \ \Tilde{d} = \frac{\log \text{det}(\mathbf{I} + \mathbf{H}/\lambda)}{ \log( 1 + Tn/\lambda)}
\]
where $\mathbf{H}$ is the neural tangent kernel matrix (NTK) \citep{zhou2020neural, ntk2018neural} and $\lambda$ is a regularization parameter. $\tilde{d}$ is the effective dimension first introduced by \citet{valko2013finite} to measure the underlying dimension of arm observed context kernel.

\begin{remark}
\textbf{Improvement by $\sqrt{\log T}$}. It is easy to observe that \sysn achieves $\sqrt{T \log T}$, improving by a multiplicative factor of $\sqrt{\log T}$ over existing works. Note that this regret bound has been around since \citep{2011improved}. We believe that this improvement provides a significant step to push the boundary forward.
\end{remark}

\begin{remark}
\textbf{Removing $d$ or $\tilde{d}$}. In the regret bound of \sysn, it does not have  $d$ or $\tilde{d}$. When input dimension is large (e.g., $d \geq T$), it may cause a considerable amount of error. The effective dimension $\tilde{d}$ may also encounter this situation when the arm context kernel matrix is very large.
\end{remark}

The analysis approach of \sysn is distinct from existing works. Clustering of bandits is based on the upper confidence bound of ridge regression \citep{2011improved}, and neural bandits \citep{zhou2020neural,zhang2020neural} use the similar method where they conduct ridge regression on the gradient space. In contrast, our analysis is based on the convergence and generalization bound of meta-learner and user-learner built on recent advances in over-parameterized networks \citep{allen2019convergence, cao2019generalization}.  The design of \sysn and new proof method are critical to \sysn's good properties.
We will provide more details in Section \ref{sec:detail}.

\section{Main Proofs} \label{sec:detail}

In this section, we provide the main lemmas in the proof of Theorem \ref{theorem:main} and explain the intuition behind them, including the convergence and an upper confidence bound (generalization bound) for $f(\cdot; \Theta)$ (Lemma \ref{flemma:metaconvergence} and \ref{flemma:ucb}), and the inferred group guarantee (Lemma \ref{flemma:groupgurantee}).

\begin{lemma} [Theorem 1 in \citep{allen2019convergence}] \label{flemma:userconvergence}
For any $ \delta \in (0,1), 0 < \epsilon_1 \leq 1$, $ 0< \rho \leq \calo(1/L)$. Given a user $u$ and the collected data $\calt^u_{t}$, suppose $m, \eta_1, J_1$ satisfy the conditions in Eq.(\ref{eq:conditions0}), then with probability at least $1 - \delta$, these hold for Algorithm \ref{alg:user} that:
\begin{enumerate}
    \item $\call(\calt^{u}_t; \widehat{\theta}^u_{\mu_t^u}) \leq \epsilon_1$ in $J_1 $ rounds.
    \item For any $j \in [J_1]$, $\| \theta^u_{(j)}  -  \theta^u_{(0)}   \| \leq \calo \left( \frac{ (\mu_t^u)^3}{ \rho \sqrt{m}} \log m \right)$.
\end{enumerate}
\end{lemma}

Lemma \ref{flemma:userconvergence} shows the loss convergence of user-learner which is the direct application of  Theorem 1 in \citep{allen2019convergence}], to ensure that the user leaner sufficiently exploits the current knowledge.

\begin{lemma} [Meta Convergence] \label{flemma:metaconvergence}
Given any  $ \delta \in (0, 1), 0 < \epsilon_1 \leq \epsilon_2 \leq 1, \lambda > 0$,  $\rho \in (0, \calo(\frac{1}{L}))$, suppose $m, \eta_1, \eta_2, J_1, J_2$ satisfy the conditions in Eq.(\ref{eq:conditions0})
and $\Theta_0, \theta_0^u, \forall u \in N$ are randomly initialized.
Define
\[\call_{\caln}(\Theta_{t,i})  = \frac{1}{2}  \underset{(\bx, r) \in \calt^u_{t-1} }{ \underset{u \in \caln }{\sum}} \left( f(\bx; \Theta_{t, i}) - r \right)^2, 
\]
where $\Theta_{t,i}$ is returned by Algorithm \ref{alg:meta} given $\caln$.
Then  with probability at least $1 - \delta$, these hold uniformly for Algorithms \ref{alg:main}-\ref{alg:user}:
\begin{enumerate}
    \item Given any $\caln \subseteq N$, $ \call_{\caln}(\Theta_{t,i}) \leq \epsilon_2$ in $J_2$ rounds, for any $t \in [T], i \in [k]$;
    \item For any $j \in [J_2]$, $\| \Theta_{(j)} - \Theta_{(0)}\|_2 \leq \beta_2$,
\end{enumerate}
where $\beta_2 =$
\[
 \calo \left( \frac{ \eta_2 n^{3/2} t^3   \sqrt{ \log^2m } +  t (\calo(t \log^2 m - \epsilon_2) ) \eta_1 \sqrt{\rho} \lambda n}{\eta_1 \rho \sqrt{m}\epsilon_2 }  \right). \\
\]
\end{lemma}

Lemma \ref{flemma:metaconvergence} shows the convergence of the meta-learner and demonstrates that  meta-learner can accurately learn the current behavior knowledge of a group.

\begin{lemma} [User Generalization] \label{flemma:usergeneral}
For any $\delta \in (0, 1)$, $ 0<\epsilon_1  \leq 1$, $\rho \in (0, \calo(\frac{1}{L}))$,  suppose $ m , \eta_1, J_1$ satisfy the conditions in Eq.(\ref{eq:conditions0}). Then with probability at least $1 -\delta$, for any $t \in [T]$, given $u \in N$ and  $\theta^u_{t-1}$,   it holds uniformly for Algorithm \ref{alg:user} that
\begin{equation}
\begin{aligned}
&\underset{\theta^u_{t-1} \sim  \{\widehat{\theta}_\tau^u\}_{\tau = 0}^{\mu^u_{t-1}} }{ \underset{(\bx, r) \sim \cald_u }{\bbe}}   [ |f(\bx; \theta^u_{t-1}) - r| | \tau^u_{t-1} ]  \\
\leq & \sqrt{\frac{2\epsilon_1}{\mu_t^u}} + \frac{3L}{\sqrt{2\mu^u_{t}}} + (1 + \xi_1)\sqrt{\frac{2 \log(\calo(\mu^u_t k)/\delta)}{\mu^u_t}} \\
= & \bar{U}_{u_t}
\end{aligned}
\end{equation}
where
$
\xi_1 =
2+  \mathcal{O} \left( \frac{t^4n L \log m}{ \rho \sqrt{m} } \right)  + \mathcal{O} \left(  \frac{t^5 n L^2  \log^{11/6} m}{ \rho  m^{1/6}}\right).
$
\end{lemma}

\begin{lemma} [Group Guarantee] \label{flemma:groupgurantee}
For any $\delta,  \epsilon_1 \in (0, 1)$, $\rho \in (0, \calo(\frac{1}{L}))$, suppose $m, \eta_1, J_1$ satisfy the conditions in Eq.(\ref{eq:conditions0}). Assume the groups in $N$ satisfy $\gamma$-gap (Definition \ref{def:gap}).  Then, given $\nu > 1$,  with probability at least $1 - \delta$, when 
\[
\begin{aligned}
 t \geq & \left( \frac{n 48\nu^2 (1 + \xi_1)^2}{\gamma^2( 1 + \sqrt{3n \log (n/\delta) })} \right)   \cdot
 \\  
 & \left( \log \frac{24\nu^2 (1 + \xi_1)^2}{\gamma^2} +  \frac{9 L^2 + 4 \epsilon_1}{4 (1 + \xi_1)^2} + \log (nk) - \log \delta  \right)    \\
  = & \widetilde{T} 
\end{aligned}
\]
for any $i \in [k]$, it uniformly holds for Algorithms \ref{alg:main}-\ref{alg:user} that 
\[
\widehat{\caln}_{u_t}(\bx_{t,i}) = \caln_{u_t}(\bx_{t,i}).
\]
\end{lemma}

Lemma \ref{flemma:usergeneral} shows a generalization bound for the user learner, inspired by \citep{cao2019generalization}.  Lemma \ref{flemma:groupgurantee} provides us the conditions under which the returned group is a highly-trusted relative group.

\begin{lemma} [Upper Confidence Bound / Generalization]  \label{flemma:ucb}
For any $\delta \in (0, 1), 0 < \epsilon_1 \leq \epsilon_2 \leq 1, \lambda > 0$, $\rho \in (0, \calo(\frac{1}{L}))$, suppose $m, \eta_1, \eta_2, J_1, J_2$ satisfy the conditions in Eq.(\ref{eq:conditions0}). Then, with probability at least $1- \delta$, for all $t \in [T], i \in [k]$, given $u_t$ and an arm $\bx_{t,i} \in \BX_t$, it holds uniformly for Algorithms \ref{alg:main}-\ref{alg:user} that
\begin{equation}
\begin{aligned}
&\underset{(\bx, r) \sim \cald_{u_t}}{\bbe}\left[ |f(\bx; \Theta_{t,i}) - r|  |  \{\calt^u_{t-1} \}_{u \in \widehat{\caln}_{u_t}(\bx_{t,i})}  \right] \\
\leq & \beta_2 \cdot  \| g(\bx; \Theta_{t, i}) - g(\bx; \theta_{0}^{u_t})\|_2 
+ Z_1 + \bar{U}_{u_t},
\end{aligned}
\end{equation}
\end{lemma}
where $Z_1 =$
\begin{equation}
\begin{aligned}
 & \mathcal{O} \left(  \frac{(t-1)^4 L^2  \log^{11/6} m}{ \rho  m^{1/6}}\right)   +
 (L+1)^2\sqrt{m \log m} \beta_2^{4/3} 
  \\
&+ \calo \left( L \left( \frac{ (t-1)^3 }{\rho \sqrt{m}} \log m  \right) \right) + \calo(L \beta_2)  \\ 
&+ \mathcal{O} \left(L^4  \left(\frac{ (t-1)^3 }{\rho \sqrt{m}} \log m  \right)^{4/3}   \right)
\end{aligned}
\end{equation}

Lemma \ref{flemma:ucb} provides the UCB which is the core to exploration in \sysn, containing both meta-side ($g(\bx; \Theta_{t, i})$)  and user side ($\bar{U}_{u_t}$) information. Then, we can simply derive the following lemma.

\begin{lemma} [Regret of One Round] \label{flemma:singleregretbound}
For any $\delta \in (0, 1), 0 < \epsilon_1 \leq \epsilon_2 \leq 1, \lambda >0$, $\rho \in (0, \calo(\frac{1}{L}))$, suppose $m, \eta_1, \eta_2, J_1, J_2$ satisfy the conditions in Eq.(\ref{eq:conditions0}). Then, with probability at least $1- \delta$, for $t \in [T]$, it holds uniformly for Algorithms \ref{alg:main}-\ref{alg:user} that
\begin{equation}
\begin{aligned}
&h_{u_t}(\bx_t^\ast)  -  h_{u_t}(\bx_t) \\  
\leq& 2 \left( \beta_2 \cdot  \| g(\bx_t^\ast; \Theta_{t}) - g(\bx_t^\ast; \Theta_{0})\|_2  +Z_1  +  \bar{U}_{u_t}  \right).
\end{aligned}
\end{equation}
\end{lemma}

\para{Proof of Theorem \ref{theorem:main}}: According to definition Eq.(\ref{eq:regret}), we have $R_{T} = \sum_{t=1}^T [h_{u_t}(\bx_t^\ast)  -  h_{u_t}(\bx_t) ]$. Therefore, we easily upper bound $R_T$. We leave all the proof details in Appendix.

\section{Experiments} \label{sec:exp}

In this section, we evaluate \sysn's empirical performance on four real-world datasets, compared to six strong state-of-the-art baselines. We first present the setup and then the results of experiments. 

\para{Movielens \citep{harper2015movielens}  and Yelp\footnote{\url{https://www.yelp.com/dataset}} datasets.} 
MovieLens is a recommendation dataset consisting of $25$ million reviews between $1.6 \times 10^5$ users and $6 \times 10^4$ movies. 
Yelp is a dataset released in the Yelp dataset challenge, composed of 4.7 million review entries made by $1.18$ million users to $1.57 \times 10^5$ restaurants.
For both these two datasets, we extract ratings in the reviews and build the rating matrix by selecting the top $2000$ users and top $10000$ restaurants(movies).
Then, we use the singular-value decomposition (SVD) to extract a normalized $10$-dimensional feature vector for each user and restaurant(movie).
The goal of this problem is to select the restaurants(movies) with bad ratings. Given an entry with a specific user, we generate the reward by using the user's rating stars for the restaurant(movie). If the user's rating is less than 2 stars (5 stars totally), its reward is $1$; Otherwise, its reward is $0$. 
From the user side, as these two datasets do not provide group information, we use K-means to divide users into 50 groups. 
Note that the group information is unknown to models.
Then, in each round, a user to serve $u_t$ is randomly drawn from a randomly selected group. 
For the arm pool,  we randomly choose one restaurant (movie) rated by $u_t$ with reward $1$ and randomly pick the other $9$ restaurants(movies) rated by $u_t$ with $0$ reward. Therefore, there are totally $10$ arms in each round. We conduct experiments on these two datasets, respectively.

\para{Mnist \citep{lecun1998gradient}, Notmnist datasets.} 
These two both are 10-class classification datasets to distinguish digits and letters. Following the evaluation setting of existing works \citep{zhou2020neural,valko2013finite,deshmukh2017multi}, we transform the classification into bandit problem.
Given an image $\bx \in \bbr^{d}$, it will be transformed into 10 arms, $\bx_1 = (\bx, 0, \dots, 0), \bx_2 = (0, \bx, \dots, 0), \dots, \bx_{10} = (0, 0, \dots, \mathbf{\bx}) \in \bbr^{d+9}$, matching 10 class in sequence. The reward is defined as $1$ if the index of selected arm equals $\bx$' ground-truth class; Otherwise, the reward is $0$.
In the experiments, we consider these two datasets as two groups, where each class can be thought of as a user. In each round, we randomly select a group (i.e., Mnist or Notmnist), and then we randomly choose an image from a class (user). Accordingly, the 10 arms and rewards are formed following the above methods. Therefore, we totally have 20 users and 2 groups in the experiments. The group information is unknown to all the models.

\para{Basedlines.} We compare \sysn to six strong baselines as follows:
\vspace{-0.5cm}
\begin{enumerate}
    \item  CLUB \citep{2014onlinecluster} clusters users based on the connected components in the user graph and refine the groups incrementally. When selecting arm, it uses the newly formed group parameter instead of user parameter with UCB-based exploration.
    \vspace{-0.2cm}
 \item COFIBA \citep{2016collaborative} clusters on both user and arm sides based on evolving graph, and chooses arms using a UCB-based exploration strategy;
 \vspace{-0.2cm}
\item SCLUB \citep{2019improved} improves the algorithm CLUB by allowing groups to merge and split to enhance the group representation;
    \vspace{-0.2cm}
\item LOCB \citep{ban2021local} uses the seed-based clustering and allow groups to be overlapped, and chooses the best group candidates when selecting arms;
    \vspace{-0.2cm}
\item NeuUCB-ONE \citep{zhou2020neural} uses one neural network to formulate all users and select arms via a UCB-based recommendation;
    \vspace{-0.2cm}
\item NeuUCB-IND \citep{zhou2020neural} uses one neural network to formulate one user separately (totally $N$ networks) and apply the same strategy to choose arms.
\end{enumerate}
\vspace{-0.5cm}
Since LinUCB \cite{2010contextual} and KernalUCB \cite{valko2013finite} are outperformed by the above baselines, we do not include them in comparison.

\begin{figure}[t] 
    \includegraphics[width=0.9\columnwidth]{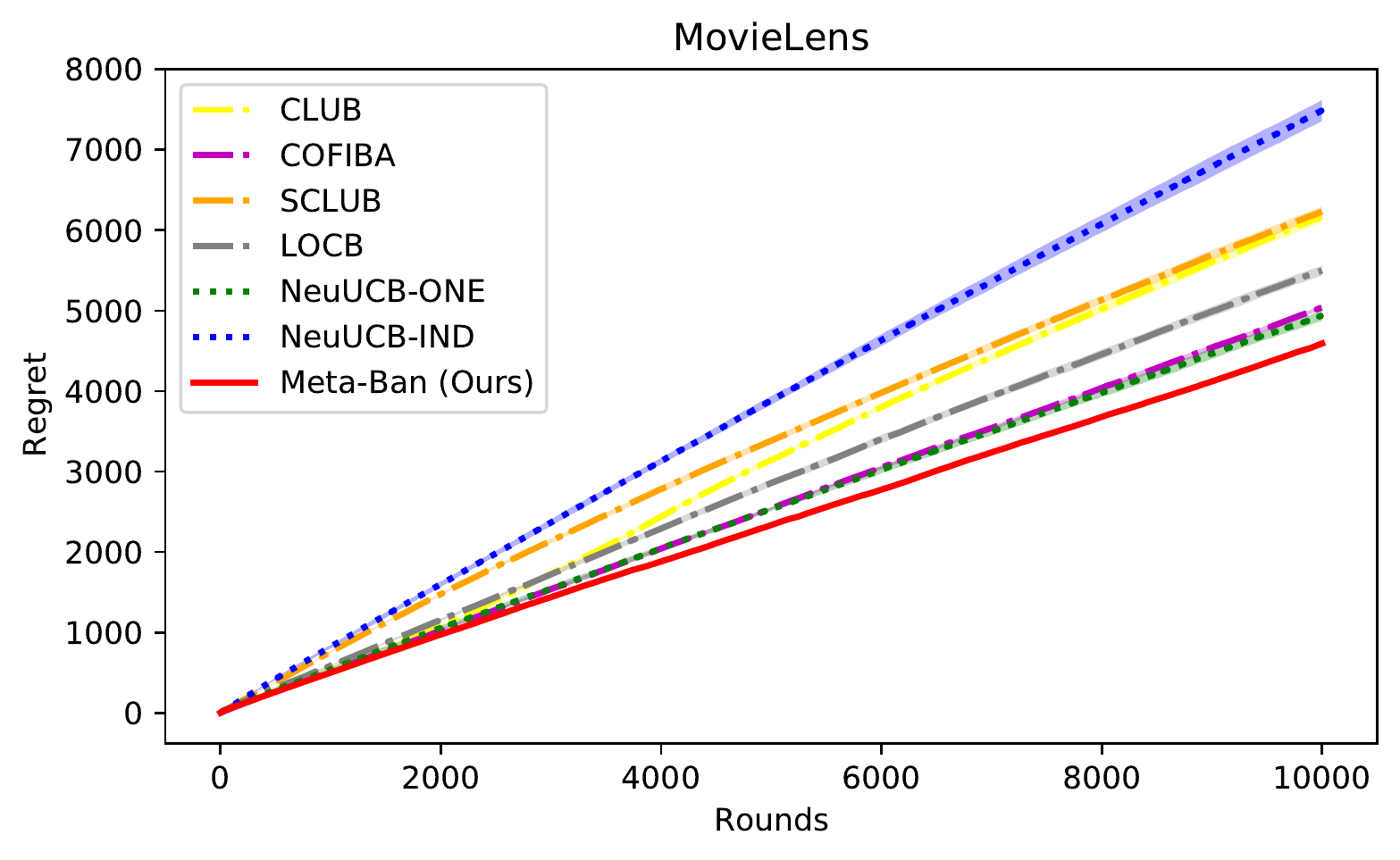}
    \centering
        \vspace{-1em}
    \caption{ Regret comparison on MovieLens.  \sysn outperforms all baselines.}
    \vspace{-2em}
       \label{fig:movie}
\end{figure}

\begin{figure}[t] 
    \includegraphics[width=0.9\columnwidth]{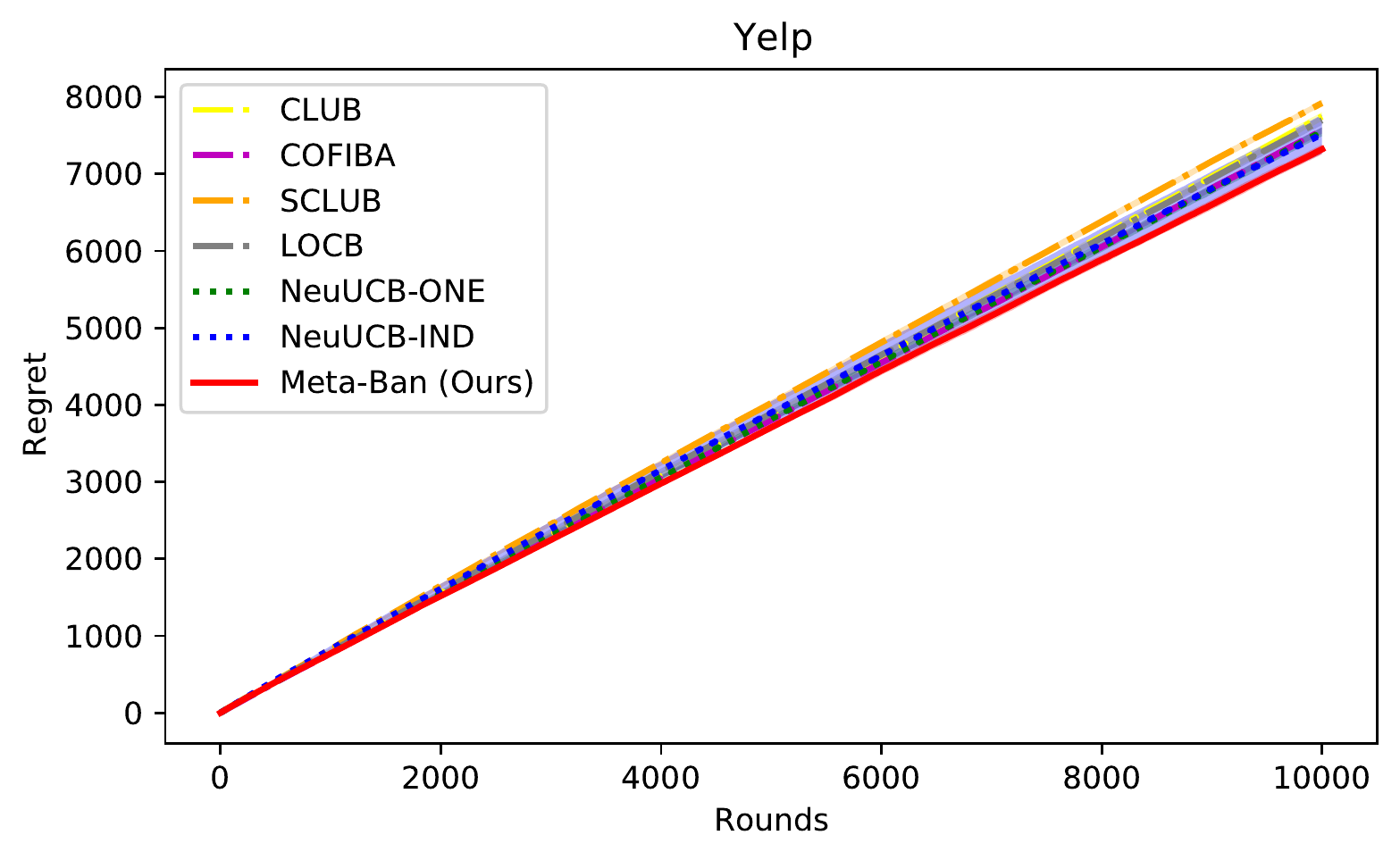}
    \centering
        \vspace{-1em}
     \caption{ Regret comparison on Yelp. \sysn outperforms all baselines.}
    \vspace{-0.5em}
       \label{fig:yelp}
\end{figure}

\para{Configurations.} For all the methods, they all have two parameters: $\lambda$ that is to tune regularization at initialization and $\alpha$ which is to adjust the UCB value. To find their best performance, we conduct the grid search for $\lambda$ and $\alpha$ over $(0.01, 0.1, 1)$ and $(0.001, 0.01, 0.1, 1)$ respectively. 
For LOCB, the number of random seeds is set as $20$ following their default setting. For $\sysn$, we set $\nu$ as $5$ and $\gamma$ as $0.4$ to tune the group set.  To compare fairly, for NeuUCB and \sysn, we use a same simple neural network with $2$ fully-connected layers and the width $m$ is set as $100$. For $\sysn$, we set $w_u = 1$ for each user. 
In the end,  we choose the best of results for the comparison and report the mean and standard deviation (shadows in figures) of $10$ runs for all methods.

\begin{figure}[t] 
    \includegraphics[width=0.9\columnwidth]{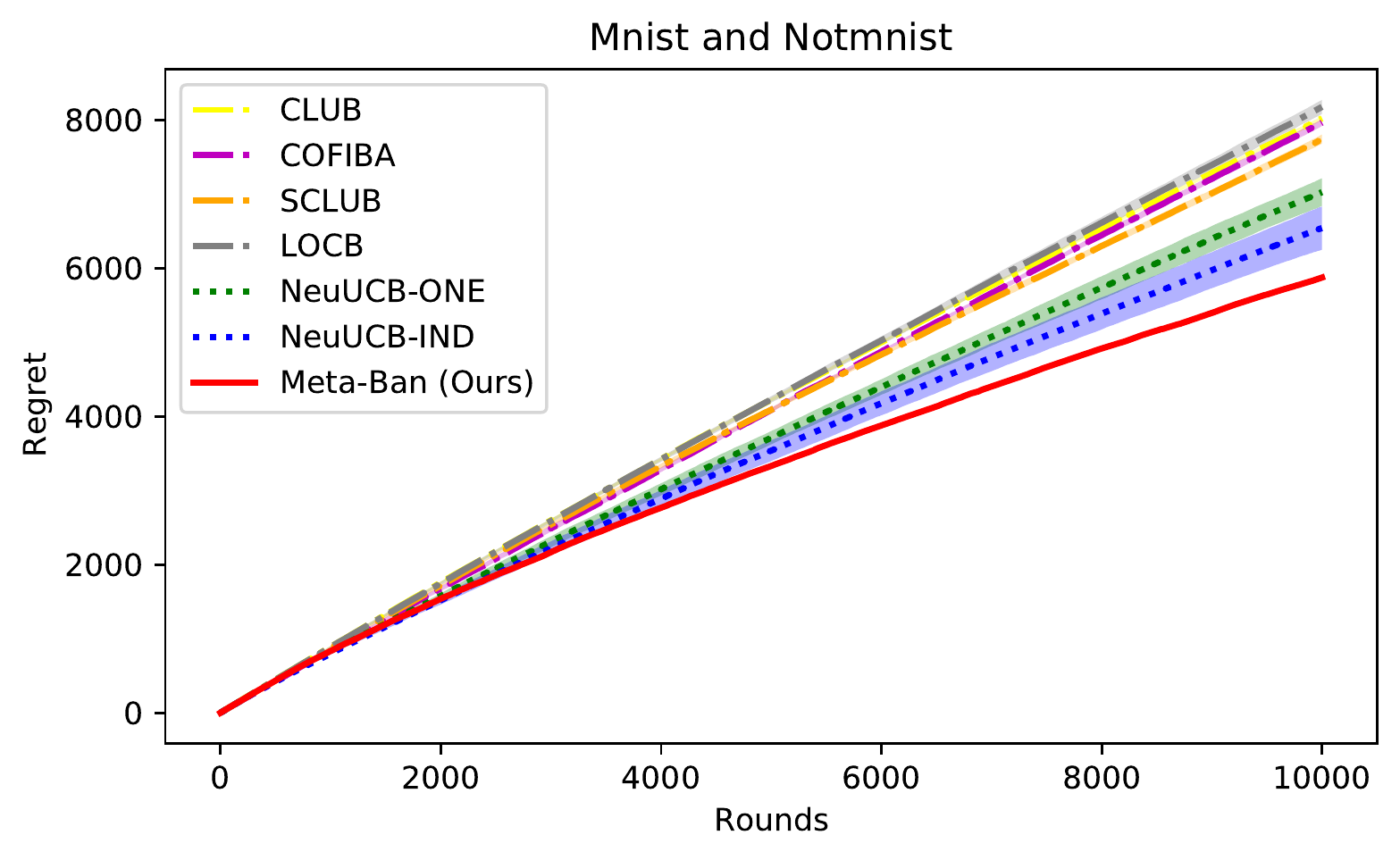}
    \centering
     \vspace{-1em}
    \caption{ Regret comparison on Mnist and Notmnist. \sysn outperforms all baselines.}
    \vspace{-2em}
       \label{fig:mnist}
\end{figure}

\para{Results.}   Figure \ref{fig:movie} and Figure \ref{fig:yelp} show the regret comparison on the two recommendation datasets in which \sysn outperforms all strong baselines. As rewards are almost linear to the arm feature vectors on these two datasets, conventional clustering of bandits (CLUB, COFIBA, SCLUB, LOCB) achieve good performance. But they still are outperformed by NeuUCB-ONE and \sysn because a simple vector cannot accurately represent a user's behavior.  Thanks to the representation power of neural networks, NeuUCB-ONE obtain good performance. However, it treats all the users as one group, neglecting the disparity among groups. In contrast, NeuUCB-IND deals with the user individually, not taking collaborative knowledge among users into account. Because \sysn uses the neural network represent users and groups while integrating collaborative information among users,  \sysn achieves the best performance.

Figure \ref{fig:mnist} reports the regret comparison on image datasets where \sysn still outperforms all baselines. As the rewards are non-linear to the arms on this combined dataset,  conventional clustering of bandits (CLUB, COFIBA, SCLUB, LOCB) perform poorly. Similarly, because \sysn can discover and leverage the group information automatically, it obtains the best performance surpassing NeuUCB-ONE and NeuUCB-IND.

Furthermore, we conducted ablation study for the group parameter $\nu$ in Appendix \ref{sec:ablation}.

\section{Conclusion} \label{sec:conclusion}

In this paper, we introduce the problem, Neural Collaborative Filtering Bandits, to incorporate collaborative effects in bandits with generic reward assumptions. Then, we propose, \sysn, to solve this paper, where a meta-learner is assigned to represent and rapidly adapt to dynamic groups. In the end, we provide a better regret upper bound for \sysn and conduct extensive experiments to evaluate its empirical performance.

\clearpage

\bibliographystyle{abbrvnat}
\bibliography{ref}

\clearpage

\appendix
\onecolumn

\section{Ablation Study for $\nu$} \label{sec:ablation}

In this section, we conduct the ablation study for the group parameter $\nu$. Here, we set $\lambda$ as a fixed value $0.4$ and change the value of $\nu$ to find the effects on \sysn's performance.

\begin{figure}[h] 
    \includegraphics[width=0.6\columnwidth]{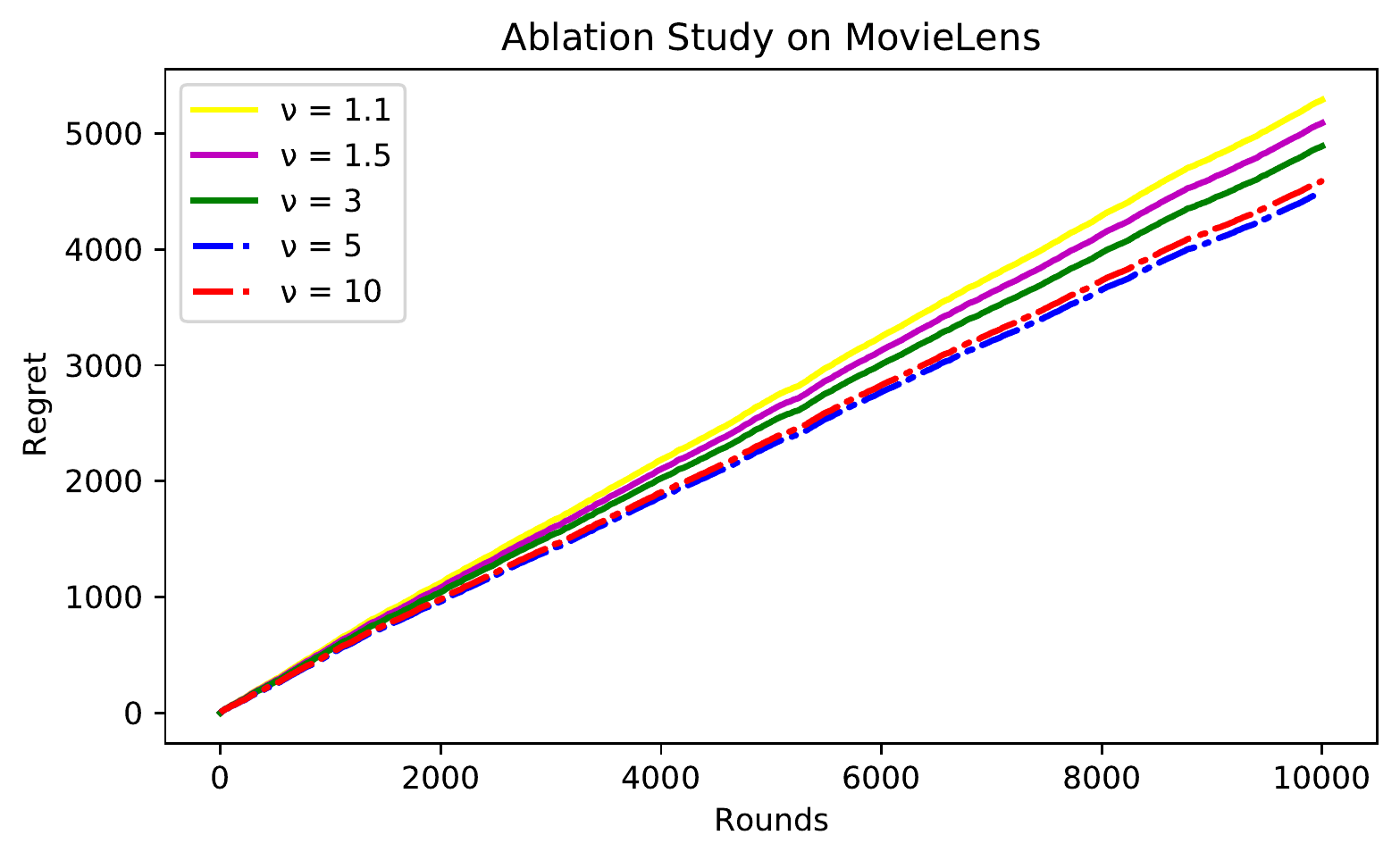}
    \centering
     \vspace{-1em}
    \caption{ Ablation study for $\nu$ on MovieLens Dataset. }
       \label{fig:ablation}
\end{figure}

Figure \ref{fig:ablation} shows the varying of performance of \sysn with respect to $\nu$. When setting $\nu = 1.1$, the exploration range of groups is very narrow. This means, in each round, the inferred group size $|\widehat{\caln}_{u_t}(\bx_{t,i})|$ tends to be small. Although the members in the inferred group $\widehat{\caln}_{u_t}(\bx_{t,i})$ is more likely to be the true member of $u_t$'s relative group, we may loss many other potential group members in the beginning phase. when setting $\nu = 5$, the exploration range of groups is wider. This indicates we have more chances to include more members in the inferred group, while this group may contain some false positives. With a larger size of group, the meta-learner $\Theta$ can exploit more information. Therefore, \sysn with $\nu = 5$ outperforms  $\nu = 1.1$. But, keep increasing $\nu$ does not mean always improve the performance, since the inferred group may consist of some non-collaborative users, bringing into noise. Therefore, in practice, we usually set $\nu$ as a relatively large number. Even we can set $\nu$ as the monotonically decreasing function with respect to $t$.

\section{Analysis in Bandits}

\begin{theorem} [Theorem \ref{theorem:main} restated]
Given the number of rounds $T$, assume that $(u_t, \cald_{u_t})$ is uniformly drawn from $\cald$, $\forall t \in [T]$. For any $\delta \in (0, 1), \rho \in (0, \calo(\frac{1}{L})),  0 < \epsilon_1 \leq \epsilon_2 \leq 1, \lambda > 0$, suppose $m, \eta_1, \eta_2, J_1, J_2$ satisfy
\begin{equation} \label{eq:conditions1}
\begin{aligned}
m &\geq  \widetilde{\Omega} \left( \max \left\{  \text{poly}(t, L, \rho^{-1}), e^{\sqrt{\log ( \calo(Tk)/ \delta)}} \right \} \right),  \ \,  \eta_1 = \boldsymbol{\Theta} \left(  \frac{\rho}{ \text{poly}(t, L) \cdot m }\right), \\
\eta_2& =  \min  \left \{  \boldsymbol{\Theta}\left( \frac{\sqrt{n}\rho}{ t^4 L^2 m} \right),    \boldsymbol{\Theta} \left( \frac{\sqrt{\rho \epsilon_2}}{ t^2 L^2 \lambda n^2} \right) \right \} ,  \ \
J_1 = \boldsymbol{\Theta} \left(  \frac{\text{poly}(t, L)}{\rho^2} \log \frac{1}{\epsilon_1}\right) \\
 J_2& = \max \left \{  \boldsymbol{\Theta}\left(  \frac{t^5(\calo(t \log^2 m) - \epsilon_2) L^2 m}{ \sqrt{n \epsilon_2}\rho }\right),  \boldsymbol{\Theta}\left( \frac{t^3 L^2 \lambda n^2 (\calo(t \log^2 m - \epsilon_2)) }{\rho \epsilon_2}   \right)\right \}.
\end{aligned}
\end{equation}
Then, with probability at least $1 - \delta$, Algorithms \ref{alg:main}-\ref{alg:user} has the following regret upper bound:
\[
\begin{aligned}
R_T \leq    2 \left( (2 \sqrt{T} -1) \sqrt{n}  + \sqrt{2 T \log (1/\delta)} \right) \cdot \left( \sqrt{2 \epsilon_1} + 3L/\sqrt{2} + (1+\xi_1)\sqrt{2 \log (\calo(Tk)/ \delta)} \right)  + \calo(1)
\end{aligned}
\]
where 
\[
 \xi_1 = 
2+  \mathcal{O} \left( \frac{t^4n L \log m}{ \rho \sqrt{m} } \right)  + \mathcal{O} \left(  \frac{t^5 n L^2  \log^{11/6} m}{ \rho  m^{1/6}}\right)
\]
For a further simplification, we have
\[
R_T  \leq  \calo(\sqrt{T} \sqrt{\log T}).
\]
\end{theorem}

\begin{proof}
\begin{equation} \label{eq1.1}
\begin{aligned}
R_T  =&  \sum_{t = 1}^ T  \left(h_{u_t}(\bx_t^\ast)  -  h_{u_t}(\bx_t) \right) \\
\underbrace{\leq}_{E_1} & \underbrace{\sum_{t = 1}^T 2 \beta_2 \cdot  \| g(\bx_t^\ast; \Theta_{t}) - g(\bx_t^\ast; \Theta_{0})\|_2 }_{I_1} + \underbrace{  \sum_{t = 1}^T \left(  \mathcal{O} \left(  \frac{(t-1)^4 L^2  \log^{11/6} m}{ \rho  m^{1/6}}\right)  
 + \mathcal{O} \left(L^4  \left(\frac{ (t-1)^3 }{\rho \sqrt{m}} \log m  \right)^{4/3}   \right)  \right) }_{I_2} \\
& +  \underbrace{ \sum_{t = 1}^T  \left( \calo \left( L \left( \frac{ (t-1)^3 }{\rho \sqrt{m}} \log m  \right) \right) + \calo(L \beta_2)   + (L+1)^2\sqrt{m \log m} \beta_2^{4/3} 
\right) } _{I_3} \\
& + 2 \underbrace{ \sum_{t = 1}^T \left( \sqrt{\frac{2\epsilon_1}{\mu_t^{u_t}}} + \frac{3L}{\sqrt{2\mu^{u_t}_{t}}} + (1 + \xi_1)\sqrt{\frac{2 \log(\calo(\mu^{u_t}_t n k)/\delta)}{\mu^{u_t}_t}}  \right) }_{I_4} \\
 \end{aligned}
\end{equation}
where $E_1$ is due to Lemma \ref{lemma:singleregretbound} and $I_2, I_3  \leq \calo(1)$ because of the choice of $m$. 

For $I_1$, according to  Lemma \ref{lemma:thetaconvergence2}, we have $\| \Theta_t - \Theta_0 \| \leq \beta_2$. Then, using Theorem 5 in \citep{allen2019convergence},  we have
\begin{equation} \label{eq1.2}
I_1 \leq  \sum_{t = 1}^T  \beta_2 \cdot \calo \left( \sqrt{\log m}  \beta_2^{1/3}L^3 \| g(\bx_t^\ast; \Theta_{0})  \|_2  \right)  \underbrace{\leq}_{E_2} \beta_2 \cdot \calo \left( \sqrt{\log m}  \beta_2^{1/3}L^4 \right)   \underbrace{\leq}_{E_3}  1
\end{equation}
where  $E_2$ is as the Lemma \ref{lemma:initilizebound} and $E_3$ is because of the choice of $m$.

For $I_4$, first, applying Hoeffding-Azuma inequality on $\frac{1}{\sqrt{ \mu_1^{u_1}}},\frac{1}{\sqrt{\mu_2^{u_2}}}, \dots,  \frac{1}{\sqrt{\mu_T^{u_T}}}$, we have
\begin{equation}
\begin{aligned}
\sum_{t =1}^T \frac{1}{\sqrt{\mu_t^{u_t}}} & \leq  \sum_{t =1}^T \bbe[\frac{1}{\sqrt{\mu_t^{u_t}}}]   + \sqrt{2T \log (1/\delta)}\\
& \leq \sum_{t =1}^T \sqrt{ \frac{ n}{t}} + \sqrt{2 T \log (1/\delta)} \\
& \leq  (2 \sqrt{T} -1) \sqrt{n}  + \sqrt{2T \log (1/\delta)} 
\end{aligned}
\end{equation}
where $\bbe[\mu_t^{u_t}] = \frac{t}{n}$  and the second inequality is because of $\sum_{t=1}^T \frac{1}{\sqrt{t}}  \leq  \int_{1}^T   \frac{1}{\sqrt{t}}  \ dx  +1 =  2 \sqrt{T} -1$ \citep{chlebus2009approximate}.
Therefore, we have
\begin{equation} \label{eq1.3}
I_4 \leq 2 \left( (2 \sqrt{T} -1) \sqrt{n}  + \sqrt{2 T \log (1/\delta)} \right) \cdot \left( \sqrt{2 \epsilon_1} + 3L/\sqrt{2} + (1+\xi_1)\sqrt{2 \log (\calo(Tk)/ \delta)} \right)
\end{equation}

Combining Eq.(\ref{eq1.1}), Eq.(\ref{eq1.2}),  and Eq.(\ref{eq1.3}) completes the proof.
Because $\xi_1 \leq \calo(2)$, we have the simplified version.
\end{proof}

\begin{lemma} [Lemma \ref{flemma:singleregretbound} restated]  \label{lemma:singleregretbound}
For any $\delta \in (0, 1), \rho \in (0, \calo(\frac{1}{L})), 0 < \epsilon_1 \leq \epsilon_2 \leq 1, \lambda > 0$, suppose $m, \eta_1, \eta_2, J_1, J_2$ satisfy the conditions in Eq.(\ref{eq:conditions1}). Then, with probability at least $1- \delta$, for $t \in [T]$, it holds uniformly for Algorithm \ref{alg:main} that
\begin{equation}
\begin{aligned}
&h_{u_t}(\bx_t^\ast)  -  h_{u_t}(\bx_t) \\  
\leq& 2\beta_2 \cdot  \| g(\bx_t^\ast; \Theta_{t}) - g(\bx_t^\ast; \Theta_{0})\|_2  + 2 Z_1 \\
& + 2 \sqrt{\frac{2\epsilon_1}{\mu_t^{u_t}}} + \frac{6L}{\sqrt{2\mu^{u_t}_{t}}} + 2 (1 + \xi_1)\sqrt{\frac{2 \log(\calo(\mu^{u_t}_t n k)/\delta)}{\mu^{u_t}_t}}.
\end{aligned}
\end{equation}
\end{lemma}
\begin{proof}

Let $\bx_t^\ast = \arg_{\bx_{t,i} \in \BX_t} \max h_{u_t}(\bx_{t, i}) $  For the regret in round $t$, we have 
\begin{equation}
    \begin{aligned}
    R_t & = \bbe[r_t^\ast  - r_t | u_t, \BX_t] \\
     & = h_{u_t}(\bx_t^\ast) - h_{u_t}(\bx_t) \\
    \end{aligned}
\end{equation}
where $\bx_t$ is the arm selected by Algorithm \ref{alg:main} in round $t$.

\begin{equation}
\begin{aligned}
R_t  = &h_{u_t}(\bx_t^\ast)  -  h_{u_t}(\bx_t)  \\
 = &  \underbrace{h_{u_t}(\bx_t^\ast)  - f(\bx_t^\ast; \theta_{t-1}^{u_t})}_{I_1}   + f(\bx_t^\ast; \theta_{t-1}^{u_t}) -      h_{u_t}(\bx_t) \\
 \leq & \underbrace{\text{UCB}_{\theta_{t-1}^{u_t}}(\bx_t^\ast)}_{I_1} +  f(\bx_t^\ast; \theta_{t-1}^{u_t}) -      h_{u_t}(\bx_t) \\
 = & \text{UCB}_{\theta_{t-1}^{u_t}}(\bx_t^\ast) +f(\bx_t^\ast; \theta_{t-1}^{u_t}) -  f(\bx_t^\ast; \Theta_{t}) | \widehat{\mathcal{N}}^{u_t}_t(\bx_t^\ast)  
     + f(\bx_t^\ast; \Theta_{t} ) | \widehat{\mathcal{N}}^{u_t}_t(\bx_t^\ast)  - h_{u_t}(\bx_t) \\
     =&  \text{UCB}_{\theta_{t-1}^{u_t}}(\bx_t^\ast) + \underbrace{\left |     f(\bx_t^\ast; \theta_{t-1}^{u_t}) -  f(\bx_t^\ast; \Theta_{t}) | \widehat{\mathcal{N}}^{u_t}_t(\bx_t^\ast)       \right |}_{I_2}  +   f(\bx_t^\ast; \Theta_{t}) | \widehat{\mathcal{N}}^{u_t}_t(\bx_t^\ast)  - h_{u_t}(\bx_t)  \\
     \leq & \text{UCB}_{\theta_{t-1}^{u_t}}(\bx_t^\ast)  +  \underbrace{ \text{UCB}_{\Theta - \theta}(\bx_t^\ast)}_{I_2} +  f(\bx_t^\ast; \Theta_{t}) | \widehat{\mathcal{N}}^{u_t}_t(\bx_t^\ast)  - h_{u_t}(\bx_t) \\
     \underbrace{\leq}_{E_1} & \text{UCB}_{\theta_{t-1}^{u_t}}(\bx_t)  +  \text{UCB}_{\Theta - \theta}(\bx_t) +  f(\bx_t; \Theta_{t}) | \widehat{\mathcal{N}}^{u_t}_t(\bx_t)  - h_{u_t}(\bx_t) \\ 
     \leq  & \text{UCB}_{\theta_{t-1}^{u_t}}(\bx_t)  +  \text{UCB}_{\Theta - \theta}(\bx_t) + | f(\bx_t; \Theta_{t}) | \widehat{\mathcal{N}}^{u_t}_t(\bx_t)  - f(\bx_t; \theta_{t-1}^{u_t}) + f(\bx_t; \theta_{t-1}^{u_t})  - h_{u_t}(\bx_t)| \\ 
     \leq  & \text{UCB}_{\theta_{t-1}^{u_t}}(\bx_t)  +  \text{UCB}_{\Theta - \theta}(\bx_t) + | f(\bx_t; \Theta_{t}) | \widehat{\mathcal{N}}^{u_t}_t(\bx_t)  - f(\bx_t; \theta_{t-1}^{u_t})| + | f(\bx_t; \theta_{t-1}^{u_t})  - h_{u_t}(\bx_t)| \\
     \leq & \text{UCB}_{\theta_{t-1}^{u_t}}(\bx_t)  +  \text{UCB}_{\Theta - \theta}(\bx_t)  +  \underbrace{\text{UCB}_{\theta_{t-1}^{u_t}}(\bx_t)  +  \text{UCB}_{\Theta - \theta}(\bx_t)}_{I_3}\\
     \leq & 2 ( \text{UCB}_{\theta_{t-1}^{u_t}}(\bx_t)  +  \text{UCB}_{\Theta - \theta}(\bx_t)) 
\end{aligned}
\end{equation}
where $I_1$ is because: As $(\bx^\ast,  r^\ast) \sim \cald_{u_t}$ and $\theta^{u_t }_{t-1} \sim  \{ \widehat{\theta}^{u_t}_\tau \}_{\tau = 0}^{\mu^{u_t}_{t-1}}$, based on Lemma \ref{lemma:genesingle}, we have
\begin{equation}\label{eq3}
\begin{aligned}
h_{u_t}(\bx_t^\ast)  - f(\bx_t^\ast; \theta_{t-1}^{u_t}) &\leq  \underset{ \theta^{u_t }_{t-1} \sim  \{ \widehat{\theta}^{u_t}_\tau \}_{\tau = 0}^{\mu^{u_t}_{t-1}} }{ \underset{(\bx, r) \sim \cald_u }{\bbe}}   [ |f(\bx; \theta^u_{t-1}) - r| | \{\bx_\tau^{u_t}, r_\tau^{u_t} \}_{\tau = 1}^{\mu_{t-1}^{u_t}} ] \\
& \leq \sqrt{\frac{2\epsilon_1}{\mu_t^{u_t}}} + \frac{3L}{\sqrt{2\mu^{u_t}_{t}}} + (1 + \xi_1)\sqrt{\frac{2 \log(\calo(\mu^{u_t}_t n k)/\delta)}{\mu^{u_t}_t}}\\
=  \text{UCB}_{\theta_{t-1}^{u_t}}(\bx^\ast)  =  \text{UCB}_{\theta_{t-1}^{u_t}}(\bx_t) = \bar{U}_{u_t}.
\end{aligned}
\end{equation}
$I_2$ is due to the direct application of Lemma \ref{lemma:thetadifference}:
\begin{equation} \label{eq4}
\begin{aligned}
&| f(\bx_t^\ast; \theta_{t-1}^{u_t}) -  f(\bx_t^\ast; \Theta_{t}) |\widehat{\mathcal{N}}^{u_t}_t(\bx_t^\ast) |   \leq 
\beta_2 \cdot  \| g(\bx_t^\ast; \Theta_{t}) - g(\bx_t^\ast; \Theta_{0})\|_2  +
Z_1 = \text{UCB}_{\Theta - \theta}(\bx_t^\ast). 
\end{aligned}
\end{equation}

$I_3$ is due to the same reason as $I_1$ and $I_2$.
$E_1$ is because of the selection criterion of Algorithm \ref{alg:main}:
\begin{equation}
\text{UCB}_{\theta_{t-1}^{u_t}}(\bx_t^\ast)  +   \text{UCB}_{\Theta - \theta}(\bx_t^\ast) +  f(\bx_t^\ast; \Theta_{t}) | \widehat{\mathcal{N}}^{u_t}_t(\bx_t^\ast)   \leq \text{UCB}_{\theta_{t-1}^{u_t}}(\bx_t)  +  \text{UCB}_{\Theta - \theta}(\bx_t) +  f(\bx_t; \Theta_{t}) | \widehat{\mathcal{N}}^{u_t}_t(\bx_t). 
\end{equation}

Finally, combining Eq.(\ref{eq3}) and (\ref{eq4}) completes the proof.
\end{proof}

\begin{lemma} [Lemma \ref{flemma:ucb} restated]
For any $\delta \in (0, 1), \rho \in (0, \calo(\frac{1}{L})), 0 < \epsilon_1 \leq \epsilon_2 \leq 1, \lambda > 0$, suppose $m, \eta_1, \eta_2, J_1, J_2$ satisfy the conditions in Eq.(\ref{eq:conditions1}). Then, with probability at least $1- \delta$, for $t \in [T]$, given $u_t$ and any arm $\bx_{t,i}$, it holds uniformly for Algorithms \ref{alg:main}-\ref{alg:user} that
\begin{equation}
\begin{aligned}
&\underset{(\bx, r) \sim \cald_{u_t}}{\bbe}\left[ |f(\bx; \Theta_{t,i}) - r|  |  \{\calt^u_{t-1} \}_{u \in \widehat{\caln}_{u_t}(\bx_{t,i})}  \right] \\
\leq & \beta_2 \cdot  \| g(\bx; \Theta_{t, i}) - g(\bx; \theta_{0}^{u_t})\|_2 
+ Z_1 + \bar{U}_{u_t},
\end{aligned}
\end{equation}
\end{lemma}

\begin{proof}
This is a simple corollary from Lemma \ref{lemma:singleregretbound}.
\begin{equation}
\begin{aligned}
 & \underset{(\bx, r) \sim \cald_{u_t}}{\bbe}\left[ |f(\bx; \Theta_{t,i}) - r|  |  \{\calt^u_{t-1} \}_{u \in \widehat{\caln}_{u_t}(\bx_{t,i})}  \right] \\
 =  &|h_{u_t}(\bx) - f(\bx; \Theta_{t, i})|\\
 =& | h_{u_t}(\bx) -f(\bx; \theta^{u_t}_{t-1})  + f(\bx; \theta^{u_t}_{t-1}) - f(\bx_{t, i}; \Theta_{t, i})|\\
 \leq & \underbrace{| h_{u_t}(\bx) -f(\bx; \theta^{u_t}_{t-1})|}_{I_1} +  \underbrace{|f(\bx; \theta^{u_t}_{t-1}) - f(\bx_{t, i}; \Theta_{t, i})|}_{I_2} \\
 \leq & \bar{U}_{u_t} +   \beta_2 \cdot  \| g(\bx_t; \Theta_{t}) - g(\bx_t; \theta_{0}^{u_t})\|_2 + Z_1,
\end{aligned}
\end{equation}
where $I_1$ is the application of Lemma \ref{lemma:genesingle} and $I_2$ is the application of Lemma \ref{lemma:thetadifference}.
\end{proof}

\begin{lemma}  \label{lemma:thetadifference}
For any $\delta \in (0, 1), \rho \in (0, \calo(\frac{1}{L})), 0 < \epsilon_1 \leq \epsilon_2 \leq 1, \lambda > 0$, suppose $m, \eta_1, \eta_2, J_1, J_2$ satisfy the conditions in Eq.(\ref{eq:conditions1}). Then, with probability at least $1- \delta$, for $t \in [T]$, given $u \in N$, $\bx_t \in \BX_t$, and $\Theta_t$ returned by Algorithm \ref{alg:meta},  it holds uniformly for Algorithms \ref{alg:main}-\ref{alg:user} that
\begin{equation}
\begin{aligned}
| f(\bx_t; \theta_{t-1}^{u}) -  f(\bx_t; \Theta_{t}) |   \leq 
\beta_2 \cdot  \| g(\bx_t; \Theta_{t}) - g(\bx_t; \theta_{0}^{u})\|_2  + Z_1.
\end{aligned}
\end{equation}
\end{lemma}

\begin{proof}

\begin{equation} \label{eq8}
\begin{aligned}
  | f(\bx_t; \theta_{t-1}^{u}) -  f(\bx_t; \Theta_{t}) | \leq &   \underbrace{|  f_{u_t}(\bx_t; \theta_{t-1}^{u})  -   \langle g(\bx_t; \theta_{t-1}^{u}), \theta_{t-1}^u - \theta_0^u \rangle  - f(\bx_t; \theta_0^u)  | }_{I_1}  \\
& +        \underbrace{ |  \langle g(\bx_t; \theta_{t-1}^{u}), \theta_{t-1}^u - \theta_0^u \rangle  + f(\bx_t; \theta_0^u)    -   f(\bx_t; \Theta_{t})| }_{I_2}
\end{aligned}
\end{equation}
where the inequality is using Triangle inequality.
For $I_1$, based on Lemma \ref{lemma:gu},  we have
\[
I_1   \leq \mathcal{O} (w^{1/3}L^2 \sqrt{m \log(m)}) \|\theta^u_{t-1} - \theta_0^u\|_2 \leq   \mathcal{O} \left(  \frac{t^4 L^2  \log^{11/6} m}{ \rho  m^{1/6}}\right),
\]
where the second equality is based on the Lemma \ref{lemma:theorem1allenzhu} (4):  $ \|\theta^{u}_{t-1} - \theta^{u}_0\|_2 \leq \mathcal{O}\left( \frac{ (\mu^u_{t-1})^3 }{\rho \sqrt{m}} \log m  \right) \leq \mathcal{O}\left( \frac{ (t-1)^3 }{\rho \sqrt{m}} \log m  \right) = w$. 

For $I_2$, we have 
\begin{equation}
\begin{aligned}
 &|  \langle g(\bx_t; \theta_{t-1}^{u}), \theta_{t-1}^u - \theta_0^u \rangle   + f(\bx_t; \theta_0^u)  -   f(\bx_t; \Theta_{t})|  \\
\underbrace{\leq}_{E_1}  &  |\langle g(\bx_t; \theta_{t-1}^{u}), \theta_{t-1}^u - \theta_0^u \rangle - \langle g(\bx_t; \Theta_{t}), \Theta_{t} - \Theta_0 \rangle   |\\
 & + |    \langle g(\bx_t; \Theta_{t}), \Theta_{t} - \Theta_0 \rangle  + f(\bx_t; \theta_0^u)       -   f(\bx_t; \Theta_{t})| \\
\underbrace{\leq}_{E_2} &  \underbrace{ |\langle g(\bx_t; \theta_{t-1}^{u}), \theta_{t-1}^u - \theta_0^u \rangle - \langle g(\bx_t; \theta_{0}^{u}), \Theta_{t} - \Theta_0 \rangle   | }_{I_3} \\
 & + \underbrace{ |  \langle g(\bx_t; \theta_{0}^{u}), \Theta_{t} - \Theta_0 \rangle       -  \langle g(\bx_t; \Theta_{t}), \Theta_{t} - \Theta_0 \rangle|}_{I_4}  \\
 &+  \underbrace{ |    \langle g(\bx_t; \Theta_{t}), \Theta_{t} - \Theta_0 \rangle + f(\bx_t; \theta_0^u)  -    f(\bx_t; \Theta_{t})| }_{I_5}
\end{aligned}
\end{equation}
where $E_1, E_2$ use Triangle inequality. For $I_3$, we have
\begin{equation}
\begin{aligned}
 & |\langle g(\bx_t; \theta_{t-1}^{u}), \theta_{t-1}^u - \theta_0^u \rangle - \langle g(\bx_t; \theta_{0}^{u}), \Theta_{t} - \Theta_0 \rangle   | \\
 \leq & |\langle g(\bx_t; \theta_{t-1}^{u}), \theta_{t-1}^u - \theta_0^u \rangle - \langle g(\bx_t; \theta_{0}^{u}), \theta_{t-1}^u - \theta_0^u \rangle| +  | \langle g(\bx_t; \theta_{0}^{u}), \theta_{t-1}^u - \theta_0^u \rangle  - \langle g(\bx_t; \theta_{0}^{u}), \Theta_{t} - \Theta_0 \rangle | \\
 \leq & \underbrace{\| g(\bx_t; \theta_{t-1}^{u}) -  g(\bx_t; \theta_{0}^{u})  \|_2 \cdot \| \theta_{t-1}^u - \theta_0^u\|_2}_{M_1} +     \underbrace{\|g(\bx_t; \theta_{0}^{u})\|_2 \cdot \|\theta_{t-1}^u - \theta_0^u - ( \Theta_{t} - \Theta_0   ) \|_2}_{M_2}
\end{aligned}
\end{equation}
For $M_1$, we have
\begin{equation}\label{eq11}
\begin{aligned}
M_1 & \underbrace{\leq}_{E_3} \mathcal{O}\left( \frac{ (t-1)^3 }{\rho \sqrt{m}} \log m  \right) \cdot  \| g(\bx_t; \theta_{t-1}^{u}) -  g(\bx_t; \theta_{0}^{u})  \|_2\\
& \underbrace{\leq}_{E_4} \mathcal{O} \left(L^4  \left(\frac{ (t-1)^3 }{\rho \sqrt{m}} \log m  \right)^{4/3}   \right)
\end{aligned}
\end{equation}
where $E_3$ is the application of Lemma \ref{lemma:theorem1allenzhu} and $E_4$ utilizes Theorem 5 in \cite{allen2019convergence} with Lemma \ref{lemma:theorem1allenzhu}.
For $M_2$, we have
\begin{equation}
\begin{aligned}
   &\|  g(\bx_t; \Theta_{0}) \|_2 \left( \| \theta_{t-1}^u - \theta_0^u    -  (  \Theta_{t} - \Theta_0)  \|_2       \right)\\
 \leq &   \|  g(\bx_t; \Theta_{0}) \|_2 \left( \| \theta_{t-1}^u - \theta_0^u  \|_2   +  \|  \Theta_{t} - \Theta_0  \|_2       \right)\\
 \underbrace{\leq}_{E_5} &      \mathcal{O}(L)          \cdot  \left[  \mathcal{O}\left( \frac{ (t-1)^3 }{\rho \sqrt{m}} \log m  \right)  +  \beta_2 \right] \\
\end{aligned}
\end{equation}\label{eq12}
where $E_5$ use Lemma \ref{lemma:initilizebound}, \ref{lemma:theorem1allenzhu}, and \ref{lemma:thetaconvergence2}. 
Combining Eq.(\ref{eq11}) and Eq.(\ref{eq12}), we have
\begin{equation} \label{eq13}
I_3 \leq  \mathcal{O} \left(L^4  \left(\frac{ (t-1)^3 }{\rho \sqrt{m}} \log m  \right)^{4/3}   \right) + \calo \left( L \left( \frac{ (t-1)^3 }{\rho \sqrt{m}} \log m  \right) \right) + \calo(L \beta_2).
\end{equation}.
For $I_4$, we have
\begin{equation} \label{eq14}
\begin{aligned}
I_4 =  & |  \langle g(\bx_t; \Theta_{0}), \Theta_{t} - \Theta_0 \rangle       -  \langle g(\bx_t; \Theta_{t}), \Theta_{t} - \Theta_0 \rangle| \\
 \leq& \| g(\bx_t; \Theta_{t}) - g(\bx_t; \Theta_{0})\|_2 \|   \Theta_{t} - \Theta_0    \|_2 \\
 \leq & \beta_2 \cdot  \| g(\bx_t; \Theta_{t}) - g(\bx_t; \Theta_{0})\|_2 
\end{aligned}
\end{equation}
where the first inequality is because of Cauchy–Schwarz inequality and the last inequality is by Lemma \ref{lemma:thetaconvergence2}.
For $I_5$, we have
\begin{equation}\label{eq15}
I_5 = |    \langle g(\bx_t; \Theta_{t}), \Theta_{t} - \Theta_0 \rangle + f(\bx_t; \Theta_0)  -    f(\bx_t; \Theta_{t})| \\
\leq  (L+1)^2\sqrt{m \log m} \beta_2^{4/3} 
\end{equation}
where this inequality uses Lemma \ref{lemma:wanggenef} with Lemma \ref{lemma:thetaconvergence2}.

Combing Eq.(\ref{eq8}), (\ref{eq13}), (\ref{eq14}), and (\ref{eq15}) completes the proof. 
\end{proof}

\begin{lemma} [Lemma \ref{flemma:groupgurantee} restated]  \label{lemma:groupgurantee}
Assume the groups in $N$ satisfy $\gamma$-gap (Definition \ref{def:gap}).
For any $\delta,  \epsilon_1 \in (0, 1)$, suppose $m, \eta_1, J_1$ satisfy the conditions in Eq.(\ref{eq:conditions1}). Then, with probability at least $1 - \delta$, when 
\[
t \geq   \frac{  n 48\nu^2 (1 + \xi_1)^2 \left( \log \frac{24\nu^2 (1 + \xi_1)^2}{\gamma^2} +  \frac{9 L^2 + 4 \epsilon_1}{4 (1 + \xi_1)^2} + \log (nk) - \log \delta  \right)   }{\gamma^2( 1 + \sqrt{3n \log (n/\delta) })}  = \widetilde{T} 
\]
for any $i \in [k]$, it uniformly holds that 
\[
\widehat{\caln}_{u_t}(\bx_{t,i}) = \caln_{u_t}(\bx_{t,i}).
\]
\end{lemma}

\begin{proof}
Given two user $u, u' \in N$ with respect one arm $\bx_{t,i} \in \BX_t$, we have
\begin{equation}
\begin{aligned}
& |h_u( \bx_{t,i} ) - h_{u'}(\bx_{t,i})| \\
 = & |h_u (\bx_{t,i}) - f(\bx_{t,i}; \theta_{t-1}^{u} ) + f(\bx_{t,i}; \theta_{t-1}^{u} ) - f(\bx_{t,i}; \theta_{t-1}^{u'} ) +f(\bx_{t,i}; \theta_{t-1}^{u'} ) - h_u (\bx_{t,i})| \\
 \leq  & |h_u (\bx_{t,i}) - f(\bx_{t,i}; \theta_{t-1}^{u} ) | + |  f(\bx_{t,i}; \theta_{t-1}^{u} ) - f(\bx_{t,i}; \theta_{t-1}^{u'} )|  +  |f(\bx_{t,i}; \theta_{t-1}^{u'} ) - h_u (\bx_{t,i})| \\
\end{aligned}
\end{equation}
According to Lemma \ref{lemma:genesingle}, for each $u \in N$, we have
\begin{equation}
\begin{aligned}
|h_u (\bx_{t,i}) - f(\bx_{t,i}; \theta_{t-1}^{u} ) | & = \underset{\theta^u_{t-1} \sim  \{\widehat{\theta}^u_\tau \}_{\tau = 0}^{\mu^u_{t-1}} }{ \underset{(\bx, r) \sim \cald_u }{\bbe}}   [ |f(\bx; \theta^u_{t-1}) - r| ] \\
& \leq \sqrt{\frac{2\epsilon_1}{\mu_t^u}} + \frac{3L}{\sqrt{2\mu^u_{t}}} + (1 + \xi_1)\sqrt{\frac{2 \log(\calo(\mu^u_t n k)/\delta)}{\mu^u_t}}.
\end{aligned}
\end{equation}
Therefore, due to the setting of Algorithm \ref{alg:main}, i.e., $ |  f(\bx_{t,i}; \theta_{t-1}^{u} ) - f(\bx_{t,i}; \theta_{t-1}^{u'} )| \leq \frac{\nu - 1}{ \nu} \gamma $,  we have
\begin{equation}
|h_u( \bx_{t,i} ) - h_{u'}(\bx_{t,i})| \leq  \frac{\nu - 1}{ \nu} \gamma  + 2 \left( \sqrt{\frac{2\epsilon_1}{\mu_t^u}} + \frac{3L}{\sqrt{2\mu^u_{t}}} + (1 + \xi_1)\sqrt{\frac{2 \log(\calo(\mu^u_t n k)/\delta)}{\mu^u_t}} \right)
\end{equation}
Next, we need to lower bound $t$ as the following:
\begin{equation}
\begin{aligned}
 \sqrt{\frac{2\epsilon_1}{\mu_t^u}} + \frac{3L}{\sqrt{2\mu^u_{t}}} + (1 + \xi_1)\sqrt{\frac{2 \log(\calo(\mu^u_t n k)/\delta)}{\mu^u_t}} & \leq \frac{\gamma}{2\nu} \\
\left(\sqrt{\frac{2\epsilon_1}{\mu_t^u}} + \frac{3L}{\sqrt{2\mu^u_{t}}} + (1 + \xi_1)\sqrt{\frac{2 \log(\calo(\mu^u_t n k)/\delta)}{\mu^u_t}} \right)^2 & \leq \frac{\gamma^2}{4\nu^2} \\
\Rightarrow  3 \left(  \left(\sqrt{\frac{2\epsilon_1}{\mu_t^u}}\right)^2 +  \left(\frac{3L}{\sqrt{2\mu^u_{t}}} \right)^2  +  \left( (1 + \xi_1)\sqrt{\frac{2 \log(\calo(\mu^u_t n k)/\delta)}{\mu^u_t}}  \right)^2  \right)& \leq \frac{\gamma^2}{4\nu^2}
\end{aligned}
\end{equation}
By simple calculations, we have
\begin{equation}
\log \mu_t^u \leq \frac{\gamma^2 \mu_t^u }{24 \nu^2 (1 + \xi_1)^2} - \frac{9 L^2 + 4 \epsilon_1}{4 (1 + \xi_1)^2} + \log \delta - \log (nk) 
\end{equation}
Then, based on Lemme 8.1 in \cite{ban2021local}, we have
\begin{equation} \label{eq22}
\mu_t^u \geq  \frac{48\nu^2 (1 + \xi_1)^2}{\gamma^2} \left( \log \frac{24\nu^2 (1 + \xi_1)^2}{\gamma^2} +  \frac{9 L^2 + 4 \epsilon_1}{4 (1 + \xi_1)^2} + \log (nk) - \log \delta  \right) 
\end{equation}
Given the binomially distributed random variables, $x_1, x_2, \dots, x_{t}$, where for $\tau \in [t]$, $x_\tau = 1$ with probability $1/n$ and $x_\tau = 0$ with probability $1 - 1/n$. Then, we have 
\begin{equation}
\mu_t^u  = \sum_{\tau =1}^t x_\tau    \ \ \text{and} \ \ \bbe[\mu_t^u] = \frac{t}{n}.
\end{equation}
Then, apply Chernoff Bounds on the $\mu_t^u$ with probability at least $1 - \delta$, for each $u \in N$,  we have
\begin{equation} \label{eq23}
\begin{aligned}
\mu_t^u  & \leq  \left(1 + \sqrt{\frac{3n \log (n/\delta)}{t}} \right) \frac{t}{n}
\Rightarrow t \geq \frac{ n \mu_t^u  }{1 + \sqrt{3n \log (n/\delta) }}   
\end{aligned}
\end{equation}
Combining Eq.(\ref{eq22}) and Eq.(\ref{eq23}), we have: When
\[
t \geq   \frac{  n 48\nu^2 (1 + \xi_1)^2 \left( \log \frac{24\nu^2 (1 + \xi_1)^2}{\gamma^2} +  \frac{9 L^2 + 4 \epsilon_1}{4 (1 + \xi_1)^2} + \log (nk) - \log \delta  \right)   }{\gamma^2( 1 + \sqrt{3n \log (n/\delta) })}  = \widetilde{T} 
\]
it holds uniformly that:
\[
  2 \left( \sqrt{\frac{2\epsilon_1}{\mu_t^u}} + \frac{3L}{\sqrt{2\mu^u_{t}}} + (1 + \xi_1)\sqrt{\frac{2 \log(\calo(\mu^u_t n k)/\delta)}{\mu^u_t}} \right) \leq \frac{\gamma}{\nu}.
\]
This indicates
\begin{equation}
|h_u( \bx_{t,i} ) - h_{u'}(\bx_{t,i})|  \leq \gamma.
\end{equation}
The proof is completed.
\end{proof}

\section{Analysis for Meta Parameters}

\begin{lemma} [Lemma \ref{flemma:metaconvergence} restated] \label{lemma:thetaconvergence2}
Given any $\delta \in (0, 1)$,  $ 0 < \epsilon_1 \leq \epsilon_2 \leq 1, \lambda>0$,  $\rho \in (0, \calo(\frac{1}{L}))$, suppose $m, \eta_1, \eta_2, J_1, J_2$ satisfy the conditions in Eq.(\ref{eq:conditions1})
and $\Theta_0, \theta_0^u, \forall u \in N$ are randomly initialized, then   with probability at least $1 - \delta$, these hold for Algorithms \ref{alg:main}-\ref{alg:user}:
\begin{enumerate}
    \item Given any $\caln \subseteq N$, define $\call_{\caln}(\Theta_{t,i})  = \frac{1}{2}  \underset{(\bx, r) \in \calt^u_{t-1} }{ \underset{u \in \caln }{\sum}} \left( f(\bx; \Theta_{t, i}) - r \right)^2$, where $\Theta_{t,i}$ is returned by Algorithm \ref{alg:meta} given $\caln$. Then, we have
    $ \call_\caln ( \Theta_{t, i}) \leq \epsilon_2$ in $J_2$ rounds.
    \item For any $j \in [J_2]$, $\| \Theta_{(j)} - \Theta_{(0)}\|_2 \leq \calo \left( \frac{ \eta_2 n^{3/2} t^3   \sqrt{ \log^2m } +  t (\calo(t \log^2 m - \epsilon_2) ) \eta_1 \sqrt{\rho} \lambda n}{\eta_1 \rho \sqrt{m}\epsilon_2 }  \right) = \beta_2$.
\end{enumerate}
\end{lemma}

\begin{proof}

Define the \textit{sign matrix}
\begin{equation}
\text{sign}(\theta_{[i]}) = 
\begin{cases}
1   \ \ \text{if} \ \theta_{[i]} \geq 0; \\
-1  \ \ \text{if} \  \theta_{[i]} < 0
\end{cases}
\end{equation}
where $\theta_{[i]}$ is the $i$-th element in $\theta$.

For the brevity, we use $\widehat{\theta}_{t}^u$ to denote $\widehat{\theta}_{\mu^u_{t}}^u$,
For each $u \in \caln$,  we have $\calt_{t-1}^u$. Given a group $\caln$,  then recall that 
\[
\call_{\caln}  =  \sum_{u \in \caln } w_{u} \cdot \call \left(\widehat{\theta}^u_{t}  \right) + \frac{\lambda}{\sqrt{m}} \sum_{u \in \caln} \| \widehat{\theta}^u_{t} \|_1.  
\]
Then, in round $t+1$, for any $j \in [J_2]$  we have
\begin{equation}
\begin{aligned}
\Theta_{(j)}  -  \Theta_{(j-1)} & =  \eta_2 \cdot \triangledown_{ \{\widehat{\theta}_{t }^u\}_{u \in \caln}}\call_{\caln}\\
& = \eta_2 \cdot \left(  \sum_{n \in \caln} \triangledown_{\widehat{\theta}_t^u} \call  +  \frac{\lambda}{\sqrt{m}} \sum_{u \in \caln} \text{sign}( \widehat{\theta}_t^u )\right)
\end{aligned}
\end{equation}
According to Theorem 4 in \citep{allen2019convergence}, given $\Theta_{(j)}, \Theta_{(j-1)}$, we have
\begin{equation}
\begin{aligned}
\call_{\caln}(\Theta_{(j)}) \leq & \call_{\caln}(\Theta_{(j-1)}) - \langle \triangledown_{\Theta_{(j-1)}} \call_{\caln}, \Theta_{(j)} - \Theta_{(j-1)}\rangle  \\
& + \sqrt{t  \call_{\caln}(\Theta_{(j-1)})} \cdot  w^{1/3} L^2 \sqrt{m \log m}  \cdot \calo( \|  \Theta_{(j)} - \Theta_{(j-1)}  \|_2 ) + \calo (tL^2m) \| \Theta_{(j)} - \Theta_{(j-1)}  \|^2_2\\
\underbrace{\leq}_{E_1} &   \call_{\caln}(\Theta_{(j-1)})  - \eta_2 \| \sum_{n \in \caln} \triangledown_{\widehat{\theta}_t^u} \call  +  \frac{\lambda}{\sqrt{m}} \sum_{u \in \caln} \text{sign}( \widehat{\theta}_t^u ) \|_2 \| \triangledown_{\Theta_{(j-1)}} \call_\caln\|_2 +    \\
 & +  \eta_2 w^{1/3} L^2 \sqrt{tm \log m}  \| \sum_{n \in \caln} \triangledown_{\widehat{\theta}_t^u} \call +  \frac{\lambda}{\sqrt{m}} \sum_{u \in \caln} \text{sign}( \widehat{\theta}_t^u )\|_2  \sqrt{  \call_{\caln}(\Theta_{(j-1)})}  \\
& + \eta_2^2  \calo (tL^2m)  \| \sum_{n \in \caln} \triangledown_{\widehat{\theta}_t^u} \call +  \frac{\lambda}{\sqrt{m}} \sum_{u \in \caln} \text{sign}( \widehat{\theta}_t^u ) \|_2^2\\
\end{aligned}
\end{equation}

\begin{equation}
\begin{aligned}
\Rightarrow \call_{\caln}(\Theta_{(j)}) \leq &  \ \call_{\caln}(\Theta_{(j-1)})  - \eta_2 \sqrt{n} \sum_{u \in \caln} \|  \triangledown_{\widehat{\theta}_t^u} \call\|_2 \| \triangledown_{\Theta_{(j-1)}} \call_\caln\|_2 +    \\
 & +  \eta_2 w^{1/3} L^2 \sqrt{t n m \log m}   \sum_{n \in \caln} \| \triangledown_{\widehat{\theta}_t^u} \call\|_2  \sqrt{  \call_{\caln}(\Theta_{(j-1)})}    + \eta_2^2  \calo (tL^2m)   n \sum_{n \in \caln} \|  \triangledown_{\widehat{\theta}_t^u} \call\|_2^2\\
 & - \frac{\eta_2 \lambda}{\sqrt{m}} \| \triangledown_{\Theta_{(j-1)}} \call_\caln\|_2 +  \eta_2 w^{1/3} n L^2 \sqrt{t \log m} \lambda  \sqrt{  \call_{\caln}(\Theta_{(j-1)})} + \calo(2 \eta^2_2 t L^2) \lambda^2 n^2 \\
 \end{aligned}
 \end{equation}

\begin{equation}
\begin{aligned}
\Rightarrow \call_{\caln}(\Theta_{(j)})  \underbrace{\leq}_{E_2}&  \call_{\caln}(\Theta_{(j-1)})  \underbrace{ - \eta_2 \sqrt{n} \sum_{u \in \caln} \frac{\rho m}{t \mu^u_t } \sqrt{\call(\widehat{\theta}_t^u) \call_\caln(\Theta_{(j-1)})} +}_{I_1} \\
 & \underbrace{ + \eta_2 w^{1/3} L^2 m \sqrt{t \rho n \log m}   \sum_{n \in \caln} \sqrt{\call(\widehat{\theta}_t^u) \call_\caln(\Theta_{(j-1)})} + \eta_2^2 t^2 L^2 m^2 n \sum_{n \in \caln}   \call(\widehat{\theta}_t^u)}_{I_1} \\
  & \underbrace{ - \frac{\eta_2 \lambda \sqrt{\rho}}{t}  \sqrt{  \call_{\caln}(\Theta_{(j-1)})}   +  \eta_2 w^{1/3} nL^2 \sqrt{t \log m} \lambda  \sqrt{  \call_{\caln}(\Theta_{(j-1)})} + \calo(2 \eta^2_2 t L^2) \lambda^2 n^2}_{I_2}  \\
\end{aligned}
\end{equation}
where $E_1$ is because of Cauchy–Schwarz inequality inequality, $E_2$ is due to Theorem 3 in \citep{allen2019convergence}, i.e., the gradient lower bound. 
Recall that
\begin{equation} \label{eq:redefine2}
\begin{aligned}
\eta_2 =  \min  \left \{  \boldsymbol{\Theta}\left( \frac{\sqrt{n}\rho}{ t^4 L^2 m} \right),    \boldsymbol{\Theta} \left( \frac{\sqrt{\rho \epsilon_2}}{ t^2 L^2 \lambda n^2 } \right) \right \} ,  \ \   \ \    L_\caln(\Theta_0) \leq \calo(t \log^2m)\\
 J_2 = \max \left \{  \boldsymbol{\Theta}\left(  \frac{t^5(\calo(t \log^2 m) - \epsilon_2) L^2 m}{ \sqrt{n \epsilon_2}\rho }\right),  \boldsymbol{\Theta}\left( \frac{t^3 L^2 \lambda n^2(\calo(t \log^2 m - \epsilon_2)) }{\rho \epsilon_2}   \right)\right \}.
\end{aligned}
\centering
\end{equation}
Before achieving  $\call_\caln(\Theta_{(j)}) \leq \epsilon_2$,  we have, for each $u \in \caln$, $ \call(\widehat{\theta}_t^u) \leq \call_\caln(\Theta_{(j-1)})$, for $I_1$, we have
\begin{equation}
\begin{aligned}
I_1 \ \  \leq &  - \eta_2 \sqrt{n} \sum_{u \in \caln} \frac{\rho m}{t \mu^u_t } \sqrt{\call(\widehat{\theta}_t^u) \call_\caln(\Theta_{(j-1)})} + \\
 & + \eta_2 w^{1/3} L^2 m \sqrt{t \rho n \log m}   \sum_{n \in \caln} \sqrt{\call(\widehat{\theta}_t^u) \call_\caln(\Theta_{(j-1)})} +  \eta_2^2 t^2 L^2 m^2 n \sum_{n \in \caln}    \sqrt{\call(\widehat{\theta}_t^u) \call_\caln(\Theta_{(j-1)})} \\
 \leq &  -  \frac{\eta_2 n \sqrt{n} \rho m }{t^2}  \sum_{n \in \caln}  \sqrt{\call(\widehat{\theta}_t^u) \call_\caln(\Theta_{(j-1)})} \\
 &+ \left(\eta_2 w^{1/3} L^2 m \sqrt{t \rho n \log m}   + \eta_2^2 t^2 L^2 m^2 n  \right)  \sum_{n \in \caln} \sqrt{\call(\widehat{\theta}_t^u) \call_\caln(\Theta_{(j-1)})} \\
\underbrace{\leq}_{E_3} &   - \boldsymbol{\Theta} \left( \frac{\eta_2 n \sqrt{n} \rho m}{t^2}  \right) \sum_{n \in \caln} \sqrt{\call(\widehat{\theta}_t^u) \call_\caln(\Theta_{(j-1)})}   \\
\underbrace{\leq}_{E_4} &  - \boldsymbol{\Theta} \left( \frac{\eta_2 n \sqrt{n} \rho m}{t^2}  \right) \sum_{n \in \caln} \call(\widehat{\theta}_t^u)
\end{aligned}
\end{equation}
where $E_3$ is because of the choice of $\eta_2$. As $\call_{\caln}(\Theta_0) \leq\calo( t \log^2m)$, we have $\call_{\caln}(\Theta_{(j)}) \leq \epsilon_2$ in $J_\Theta$ rounds.
For $I_2$, we have 
\begin{equation}
\begin{aligned}
I_2 & \underbrace{\leq}_{E_5} - \frac{\eta_2 \lambda \sqrt{\rho}}{t}  \sqrt{ \epsilon_2  }   +  \eta_2 w^{1/3} n L^2 \sqrt{t \log m} \lambda  \sqrt{  \call_{\caln}(\Theta_{(0)})} + \calo(2 \eta^2_2 t L^2) \lambda^2 n^2  \\
& \underbrace{\leq}_{E_6} - \frac{\eta_2 \lambda \sqrt{\rho}}{t}  \sqrt{ \epsilon_2  }   +  \eta_2 w^{1/3} n L^2 \sqrt{t \log m} \lambda  \sqrt{ \calo(t \log^2m)} + \calo(2 \eta^2_2 t L^2) \lambda^2 n^2 \\
& \leq \left(  - \frac{\eta_2  \sqrt{\rho}}{t}  \sqrt{ \epsilon_2  }  +   \eta_2 w^{1/3} n L^2 \sqrt{t \log m}   \sqrt{ \calo(t \log^2m)}    + \calo(2 \eta^2_2 t L^2) \lambda n^2  \right)  \lambda \\
& \underbrace{\leq}_{E_7}  - \boldsymbol{\Theta}(\frac{\eta_2 \sqrt{\rho \epsilon_2}}{t}) \lambda 
\end{aligned}
\end{equation}
where $E_5$ is by $\call_\caln(\Theta_{(j-1)}) \geq \epsilon_2$ and $\call_\caln(\Theta_{(j-1)}) \leq \call_\caln(\Theta_{(0)})$, $E_6$ is according to Eq.(\ref{eq:redefine2}), and $E_7$ is because of the choice of $\eta_2$.

Combining above inequalities together, we have
\begin{equation}
\begin{aligned}
\call_{\caln}(\Theta_{(j)}) \leq & \call_{\caln}(\Theta_{(j-1)}) -  \boldsymbol{\Theta} \left( \frac{\eta_2 n \sqrt{n} \rho m}{t^2}  \right) \sum_{n \in \caln} \call(\widehat{\theta}_t^u)  - \boldsymbol{\Theta}(\frac{\eta_2 \sqrt{\rho \epsilon_2}}{t}) \lambda  \\
\leq &  \call_{\caln}(\Theta_{(j-1)})  - \boldsymbol{\Theta}(\frac{\eta_2 \sqrt{\rho \epsilon_2}}{t}) \lambda \\
\end{aligned}
\end{equation}
Thus, because of the choice of $J_2, \eta_2$, we have
\begin{equation}
\begin{aligned}
\call_{\caln}(\Theta_{(J_2)}) & \leq \call_{\caln}(\Theta_{(0)}) - J_2 \cdot \boldsymbol{\Theta}(\frac{\eta_2 \sqrt{\rho \epsilon_2}}{t}) \lambda \\
&\leq  \calo(t \log^2m) - J_2 \cdot \boldsymbol{\Theta}(\frac{\eta_2 \sqrt{\rho \epsilon_2}}{t}) \leq \epsilon_2.\\
\end{aligned}
\end{equation}
The proof of (1) is completed.

According to Lemma \ref{lemma:theorem1allenzhu}, For any $j \in [J_1]$, $\call(\theta^u_{(j)}) \leq (1 - \Omega(\frac{\eta \rho m}{d {\mu^u_t}^2})) \call(\theta^u_{(j-1)}) $. Therefore, for any $u \in [n]$, we have
\begin{equation} \label{eq:loss35}
\begin{aligned}
\sqrt{\call(\widehat{\theta}_t^u)} & \leq    \sum_{ j =  0 }^{J_1} \sqrt{ \call(\theta_{(j)}^u)} \leq  \calo \left( \frac{(\mu^u_t)^2}{\eta_1 \rho m}\right) \cdot \sqrt{\call(\theta_{(0)}^u) } \\
& \leq   \calo \left( \frac{(\mu^u_t)^2}{\eta_1 \rho m}\right) \cdot   \calo (\sqrt{ \mu^u_t  \log^2 m} ),
\end{aligned}
\end{equation}
where the last inequality is because of Lemma \ref{lemma:theorem1allenzhu} (3).

Second, we have 
\begin{equation} \label{eq:thetai}
\begin{aligned}
\|\Theta_{(J_2)} - \Theta_0\|_2 & \leq \sum_{j=1}^{J_2} \|\Theta_{(j)} - \Theta_{(j-1)}\|_2\\
& \leq \sum_{j=1}^{J_2} \eta_2 \| \sum_{n \in \caln} \triangledown_{\widehat{\theta}_t^u} \call  +  \frac{\lambda}{\sqrt{m}} \sum_{u \in \caln} \text{sign}( \widehat{\theta}_t^u )   \|_2  \\
&  \leq \underbrace{\sum_{j=1}^{J_2} \eta_2 \| \sum_{u \in \caln} \triangledown_{\widehat{\theta}^u_t}\call  \|_F}_{I_3}   + \frac{ J_2 \eta_2 \lambda n}{ \sqrt{m}} \\
\end{aligned}
\end{equation}

For $I_3$, we have
\begin{equation} 
\begin{aligned}
\sum_{j=1}^{J_2} \eta_2 \| \sum_{u \in \caln} \triangledown_{\widehat{\theta}^u_t}\call  \|_2   &\leq \sum_{j=1}^{J_2}  \eta_2 \sqrt{|\caln|}  \sum_{u \in  \caln}  \|  \triangledown_{\widehat{\theta}^u_t}\call  \|_2 \\
& \underbrace{\leq}_{E_8}  \sum_{j=1}^{J_2} \eta_2 \sqrt{n} \sum_{u \in  N}  \|  \triangledown_{\widehat{\theta}^u_t}\call  \|_2\\
& \underbrace{\leq}_{E_9}  \calo \sum_{j=1}^{J_2} (\eta_2)  \sqrt{n t m} \sum_{u \in N} \sqrt{  \call(\widehat{\theta}_t^u)  } \\
\end{aligned}
\end{equation}

\begin{equation} \label{eq:thetai3}
\begin{aligned}
\Rightarrow  \sum_{j=1}^{J_2} \eta_2 \| \sum_{u \in \caln} \triangledown_{\widehat{\theta}^u_t}\call  \|_2   & \underbrace{\leq}_{E_{10}} \calo  (\eta_2)  \sqrt{n t m} \sum_{u \in N} \sum_{ j =  1 }^{J_2} \sqrt{ \call(\widehat{\theta}_t^u)} \\
& \underbrace{\leq}_{E_{11}} \calo (\eta_2)  \sqrt{n t m} \cdot n \cdot \calo \left( \frac{(\mu^u_t)^2}{\eta_1 \rho m}\right) \cdot   \calo (\sqrt{ \mu^u_t  \log^2 m} )\\
& \leq \calo \left( \frac{ \eta_2 n^{3/2} t^{5/2}   \sqrt{t \log^2m }}{\eta_1 \rho \sqrt{m}}     \right)
\end{aligned}
\end{equation}
where $E_1$ is because of $ |\caln| \leq n$, $E_2$ is due to Theorem 3 in \citep{allen2019convergence}, and $E_3$ is as the result of Eq.(\ref{eq:loss35}).

Combining Eq.(\ref{eq:thetai}) and Eq.(\ref{eq:thetai3}), we have
\begin{equation}
\begin{aligned}
\|\Theta_{(J_2)} - \Theta_0\|_2 \leq   \calo \left( \frac{ \eta_2 n^{3/2} t^3   \sqrt{ \log^2m } +  J_2 \eta_2 \eta_1 \rho \lambda n}{\eta_1 \rho \sqrt{m}}  \right)\\
\leq  \calo \left(  \frac{\eta_2 n^{3/2} t^3   \sqrt{ \log^2m } +  t (\calo(t \log^2 m - \epsilon_2) ) \eta_1 \sqrt{\rho} \lambda n}{\eta_1 \rho \sqrt{m}\epsilon_2 }  \right).
\end{aligned}
\end{equation}
The proof is completed.
\end{proof}

\subsubsection{Another Form of Meta Generalization}

Here, we provide another version of Algorithm \ref{alg:meta} to update the meta-learner, as described in Algorithm \ref{alg:meta1}. Lemma \ref{lemma:metageneral} shows another generalization bound for the meta learner, which also can be thought of as a UCB for $\Theta$. However, compared to Lemma \ref{flemma:ucb}, Lemma \ref{lemma:metageneral} loses the information contained in meta gradients and user-side information. Therefore, we choose Lemma \ref{flemma:ucb} as the UCB of \sysn.

\begin{algorithm}[h]
\caption{SGD\_Meta ($\caln$) }\label{alg:meta1}
\begin{algorithmic}[1]
\STATE $\Theta_{(0)} =  \Theta_{0}$ (or  $\Theta_{t-1}$) \ \
\STATE $\calt^{\caln}_{t-1} = \emptyset$
\FOR{ $u \in \caln$}
\STATE Collect $ \mathcal{T}^{u}_{t-1}$ 
\STATE $\calt^{\caln}_{t-1} = \calt^{\caln}_{t-1} \cup  \mathcal{T}^{u}_{t-1}$
\ENDFOR
\FOR{$ j = 1, 2, \dots, | \calt^{\caln}_{t-1}| $}
\STATE Sequentially choose $ (\bx_j, r_j)  \in  \calt^{\caln}_{t-1} $ and corresponding $u_j$
\STATE $\call_j\left( \widehat{\theta}^{u_j}_{t-1}  \right) = \frac{1}{2} ( f(\bx_j;\widehat{\theta}^{u_j}_{t-1}) - r_j)^2  $
\STATE $\Theta_{(j)} = \Theta_{(j-1)} -  \eta_2 \triangledown_{\widehat{\theta}^{u_j}_{t-1}  }\call_j$
\ENDFOR
\STATE \textbf{Return:} Choose $\Theta_t$ uniformly from $\{\Theta_{(0)},  \dots, \Theta_{(| \calt^{\caln}|-1)}\}$ 
\end{algorithmic}
\end{algorithm}

\begin{lemma} \label{lemma:metageneral}
For any $\delta \in (0, 1), \rho \in (0, \calo(\frac{1}{L}))$, suppose $ 0<\epsilon_1, \epsilon_2  \leq 1$, and $m, \eta_1, \eta_2, J_1, J_2$ satisfy the conditions in Eq.(\ref{eq:conditions1}). Then with probability at least $1 -\delta$, for any $t \in [T]$, given a group $\caln$ and it's historical data $\calt^\caln_{t-1}$ where $\bar{t} = |\calt^\caln_{t-1}|$, it holds uniformly for Algorithm \ref{alg:meta1} that
\[
\underset{ \Theta_{t} \sim \{ \Theta_{(\tau-1)}\}_{\tau =1}^{\bar{t}}   }{\underset{(\bx, r) \sim \cald_u}{\underset{  (u, \cald_{u}) \sim \cald}{\bbe}}}\left[ |f(\bx; \Theta_{t}) - r|  | \calt^\caln_{t-1}  \right] \leq \sqrt{ \frac{2\epsilon_2}{\bar{t}}} + \frac{3L}{\sqrt{2\bar{t}}}  + (1+ \xi_2)\sqrt{\frac{ 2 \log(\calo(\bar{t}k)/\delta)}{\bar{t}}},
\]
where $\Theta_{t}$ is uniformly drawn from $\{ \Theta_{(\tau-1)}\}_{\tau =1}^{\bar{t}}$.
\end{lemma}

\begin{proof}
Let $\bar{t} = |\calt^\caln_{t-1}|$ and use $\{\bx_\tau, r_\tau\}_{\tau=1}^{\bar{t}}$ to represent $\calt^\caln_{t-1}$.

According to Lemma \ref{lemma:fthetabound2}, with probability at least $1 - \delta$, given any $\|\bx\|_2 = 1, r \leq 1$, for any $\tau \in [\bar{t}]$, we have
\begin{equation}
| f(\bx; \Theta_{(\tau)}) - r| \leq 1 + \xi_2.
\end{equation}
Then, define
\begin{equation}
V_\tau =  \underset{(\bx, r) \sim \cald_u}{\underset{  (u, \cald_{u}) \sim \cald}{\bbe}}\left[ |f(\bx; \Theta_{(\tau-1)}) - r| \right]  - | f(\bx_\tau; \Theta_{(\tau-1)})  - r_\tau|
\end{equation}
Then, we have
\begin{equation}
\bbe[V_{\tau} | \mathbf{F}_\tau] =    \underset{(\bx, r) \sim \cald_u}{\underset{  (u, \cald_{u}) \sim \cald}{\bbe}}\left[ |f(\bx; \Theta_{(\tau-1)}) - r| \right]  - \bbe \left[ | f(\bx_\tau; \Theta_{(\tau-1)})  - r_\tau| \right] = 0
\end{equation}
where $\mathbf{F}_\tau$ represents the $\sigma$-algebra generated by $\{\bx_{\tau'}, r_{\tau'} \}_{\tau' =1}^\tau$.
Therefore, according to Lemma 1 in \citep{cesa2004generalization}, applying Hoeffding-Azuma inequality to the bounded variables $V_1, \dots, V_t$, we have
\begin{equation}
\begin{aligned}
\frac{1}{\bar{t}} \sum_{\tau = 1}^{\bar{t}}  \underset{(\bx, r) \sim \cald_u}{\underset{  (u, \cald_{u}) \sim \cald}{\bbe}}\left[ |f(\bx; \Theta_{(\tau-1)}) - r| \right]  
\leq  \frac{1}{\bar{t}}\sum_{\tau = 1}^{\bar{t}} | f(\bx_\tau; \Theta_{(\tau-1)})  - r_\tau| + (1+ \xi_2)\sqrt{\frac{ 2 \log(1/\delta)}{\bar{t}}}.
\end{aligned}
\end{equation}
Accoding to Algorithm \ref{alg:meta1}, we have
\begin{equation}
 \underset{ \Theta_{t} \sim \{ \Theta_{(\tau-1)}\}_{\tau=1}^{\bar{t}}   }{\underset{(\bx, r) \sim \cald_u}{\underset{  (u, \cald_{u}) \sim \cald}{\bbe}}}\left[ |f(\bx; \Theta_{t}) - r| \right]   =  \frac{1}{\bar{t}} \sum_{\tau = 1}^{\bar{t}}  \underset{(\bx, r) \sim \cald_u}{\underset{  (u, \cald_{u}) \sim \cald}{\bbe}}\left[ |f(\bx;  \Theta_{(\tau-1)}) - r| \right]  
\end{equation}
Therefore, we have
\begin{equation} \label{eq:g1}
\underset{ \Theta_{t} \sim \{ \Theta_{(\tau-1)}\}_{\tau =1}^{\bar{t}}   }{\underset{(\bx, r) \sim \cald_u}{\underset{  (u, \cald_{u}) \sim \cald}{\bbe}}}\left[ |f(\bx; \Theta_{t}) - r| \right] \leq  \underbrace{ \frac{1}{\bar{t}}\sum_{\tau = 1}^{\bar{t}} | f(\bx_\tau; \Theta_{(\tau-1)})  - r_\tau|}_{I_1} + (1+ \xi_2)\sqrt{\frac{ 2 \log(1/\delta)}{ \bar{t}}}.
\end{equation}
For $I_1$, Applying the Lemma \ref{lemma:caogenelimetab}, for any $\Theta' \in \bbr^p$ satisfying $\|\Theta' - \Theta_0\|_2 \leq \beta_2$, we have
\begin{equation} \label{eq:g2}
\frac{1}{\bar{t}}\sum_{\tau = 1}^{\bar{t}} | f(\bx_\tau; \Theta_{(\tau-1)})  - r_\tau| \leq \underbrace{ \frac{1}{\bar{t}}\sum_{\tau = 1}^{\bar{t}}  | f(\bx_\tau; \Theta')  - r_\tau| }_{I_2} + \frac{3L}{\sqrt{2 \bar{t}}}
\end{equation}
For $I_2$, we have
\begin{equation}\label{eq:g3}
\frac{1}{\bar{t}}\sum_{\tau = 1}^{\bar{t}}  | f(\bx_\tau; \Theta')  - r_\tau| \leq \frac{1}{\bar{t}}\sqrt{\bar{t}} \sqrt{  \sum_{\tau = 1}^{\bar{t}}  ( f(\bx_\tau; \Theta')  - r_\tau)^2}
\leq \frac{1}{\sqrt{\bar{t}}} \sqrt{2\epsilon_2}.
\end{equation}
where the last inequality is according to \ref{lemma:thetaconvergence2}: there exists $ \Theta'$ satisfying  $ \| \Theta' - \Theta_0\|_2 \leq \beta_2 $, such that $\frac{1}{2} \sum_{\tau=1}^{\bar{t}} ( f(\bx_\tau; \Theta') - r_\tau)^2 \leq \epsilon_2$.

In the end, combining Eq.(\ref{eq:g1}), Eq.(\ref{eq:g2}), and Eq.(\ref{eq:g3}), we have
\begin{equation}
\underset{ \Theta_{t} \sim \{ \Theta_{(\tau-1)}\}_{\tau =1}^{\bar{t}}   }{\underset{(\bx, r) \sim \cald_u}{\underset{  (u, \cald_{u}) \sim \cald}{\bbe}}}\left[ |f(\bx; \Theta_{\tau}) - r| \right] \leq \sqrt{ \frac{2\epsilon_2}{\bar{t}}} + \frac{3L}{\sqrt{2 \bar{t}}}  + (1+ \xi_2)\sqrt{\frac{ 2 \log(1/\delta)}{\bar{t}}}.
\end{equation}
Apply the union bound for $\tau \in [\bar{t}], i \in [k]$ and the proof is completed.
\end{proof}

\begin{lemma}\label{lemma:fthetabound2}
For any $\delta \in (0, 1), \rho \in (0, \calo(\frac{1}{L}))$, $0 < \epsilon_1 \leq \epsilon_2 < 0, \lambda > 0$, suppose $m, \eta_1, \eta_2, J_1, J_2$ satisfy the conditions in Eq.(\ref{eq:conditions1}). Then, with probability at least $1-\delta$, for any $j \in [\bar{t}]$ we have
\[
| f(\bx; \Theta_{(j)})| \leq \left( 2 +  \calo(L\beta_2) + \mathcal{O}((L+1)^2 \sqrt{m \log m} \beta_2^{4/3})\right) = \xi_2
\]

\end{lemma}

\begin{proof}
Based on Lemma \ref{lemma:wanggenef}, 
Then, we have
\begin{equation}
\begin{aligned}
\Rightarrow  
 | f(\bx; \Theta_{(j)})  |   \leq    & \underbrace{| f(\bx; \Theta_0)  |}_{I_1} +  \underbrace{ |  \langle \triangledown_{\Theta}f(\bx; \Theta_{0}),   \Theta_{(j)} - \Theta_{0} \rangle |}_{I_2} +     \mathcal{O}(w^{1/3}(L+1)^2 \sqrt{m \log m}) \| \Theta_{(j)} - \Theta_{0}  \|_2 \\
\leq  & \underbrace{ 2\|\bx\|_2}_{I_1} + \underbrace{ \|\triangledown_{\Theta}f(\bx; \Theta_{0})\|_2 \|   \Theta_{(j)} - \Theta_{0}   \|_2}_{I_2}  + \mathcal{O}(w^{1/3}(L+1)^2 \sqrt{m \log m}) \| \Theta_{(j)} - \Theta_{0}  \|_2 \\ 
\underbrace{\leq}_{E_3} & 2  + \calo(L) \cdot \beta_2 +  \mathcal{O}((L+1)^2 \sqrt{m \log m})\beta_2^{4/3}
\end{aligned}
\end{equation}
where $I_1$ is the applying of Lemma 7.3 in \citep{allen2019convergence}, $I_2$ is by Cauchy–Schwarz inequality, and $E_3$ is due to Lemma \ref{lemma:thetaconvergence2}.
\end{proof}

\begin{lemma}[\citep{wang2020globalcon}] \label{lemma:wanggenef}
Suppose $m$ satisfies the condition2 in  Eq.(\ref{eq:conditions1}), if
\[
     \Omega( m^{-3/2}L^{-3/2} [\log (TkL^2/\delta)]^{3/2} ) \leq \nu \leq \mathcal{O}((L+1)^{-6}\sqrt{m}).
\]
then with probability at least $1 - \delta$, for all $\Theta, \Theta'$ satisfying $\| \Theta - \Theta_0\|_2 \leq \nu$ and $\|\Theta' - \Theta_0\|_2 \leq \nu$, $\bx \in \bbr^d$,  $\|\bx\|_2 = 1$,  we have
\[
|f(\bx; \Theta) - f(\bx; \Theta') - \langle \triangledown_{\Theta}f(\bx; \Theta),   \Theta' - \Theta \rangle| \leq \mathcal{O}( \nu^{4/3} (L+1)^2 \sqrt{m \log m} ).
\]
\end{lemma}

\begin{lemma} \label{lemma:caogenelimetab}
For any $\delta \in (0, 1)$, suppose
\[
m > \tilde{\mathcal{O}}\left(  \text{poly} (T, n, \delta^{-1}, L) \cdot  \log (1/\delta) \cdot e^{\sqrt{ \log 1/\delta}} \right), \ \ \nu = \boldsymbol{\Theta}(\bar{t}^6/\delta^2).  
\]
Then, with probability at least $1 - \delta$, set $\eta_2 =  \Theta( \frac{ \nu}{\sqrt{2} \bar{t} m})$, for any $\Theta' \in \bbr^p$ satisfying $\|\Theta' - \Theta_0\|_2 \leq \beta_2$ ,  such that
\[
\begin{aligned}
 \sum_{\tau=1}^{\bar{t}} | f(\bx_{\tau}; \Theta_{(j)} - r_\tau|  & \leq \sum_{\tau =1 }^{\bar{t}} | f(\bx_{\tau}; \Theta') - r_\tau| + \frac{3L\sqrt{ \bar{t}} }{\sqrt{2}} \\
\end{aligned}
\]
\end{lemma}
\begin{proof}
Then, the proof is a direct application of Lemma 4.3 in \citep{cao2019generalization} by setting the loss as $L_\tau(\widehat{\Theta}_{\tau}) = | f(\bx_{\tau}; \widehat{\Theta}_{\tau}) - r_\tau|  $, $R = \beta_2 \sqrt{m} ,  \epsilon = \frac{LR}{ \sqrt{2\nu \bar{t}}}$,  and $ \nu = R^2$.

\end{proof}

\section{Analysis for User Parameters}

\begin{lemma} [Lemma \ref{flemma:usergeneral} restated]   \label{lemma:genesingle}
For any $\delta \in (0, 1), \rho \in (0, \calo(\frac{1}{L}))$, suppose $ 0<\epsilon_1 \leq 1$ and $m , \eta_1, J_1$ satisfy the conditions in Eq.(\ref{eq:conditions1}). Then with probability at least $1 -\delta$, for any $t \in [T]$, given $u \in N$,  it holds uniformly for Algorithms \ref{alg:main}-\ref{alg:user} that
\[
\underset{\theta^u_{t-1} \sim  \{ \widehat{\theta}_\tau^u \}_{\tau = 0}^{\mu^u_{t-1}} }{ \underset{(\bx, r) \sim \cald_u }{\bbe}}   [ |f(\bx; \theta^u_{t-1}) - r| \calt_{t-1}^u ] \leq \sqrt{\frac{2\epsilon_1}{\mu_t^u}} + \frac{3L}{\sqrt{2\mu^u_{t}}} + (1 + \xi_1)\sqrt{\frac{2 \log(\calo(\mu^u_t k)/\delta)}{\mu^u_t}}.
\]
where $\theta^u_{t-1} $ is uniformly drawn from $\{ \widehat{\theta}_\tau^u \}_{\tau = 0}^{\mu^u_{t-1}}$.
\end{lemma}

\begin{proof}
According to Lemma \ref{lemma:xi1},  with probability at least $1 - \delta$, given any $\|\bx\|_2 = 1, r \leq 1$, for any $\tau \in [\mu_{t}^u]$, we have
\[
| f(\bx; \widehat{\theta}_{\tau-1}^u) - r| \leq \xi_1+1.
\]

First, given a user $u \in N$, we have $u$'s collected data $ \calt_{t}^u$. Then,
for each $\tau \in [\mu_t^u]$,  define 
\begin{equation}
V_\tau = \bbe_{(\bx, r) \sim \cald_u}[ |f(\bx; \widehat{\theta}^u_{\tau-1}) - r| ] - | f(\bx_{\tau}; \widehat{\theta}^u_{\tau-1}) - r_\tau|.
\end{equation}
Then, we have
\[
\bbe[V_\tau|\mathbf{F}_\tau] =  \bbe_{(\bx, r) \sim \cald_u}[ |f(\bx; \widehat{\theta}^u_{\tau-1}) - r| ] - \bbe[| f(\bx_{\tau}; \widehat{\theta}^u_{\tau-1}) - r_\tau| | \mathbf{F}_\tau] = 0
\]
where $\mathbf{F}_\tau$ denotes the $\sigma$-algebra generated by $\calt_{\tau-1}^u $. Thus, we have the following form:
\begin{equation}
\frac{1}{\mu_t^u} \sum_{\tau=1}^{\mu_t^u} V_\tau = \frac{1}{\mu_t^u} \sum_{\tau=1}^{\mu_t^u}  \bbe_{(\bx, r) \sim \cald_u}[ |f(\bx; \widehat{\theta}^u_{\tau-1}) - r| ] -  \frac{1}{\mu_t^u} \sum_{\tau=1}^{\mu_t^u} | f(\bx_{\tau}; \widehat{\theta}^u_{\tau-1}) - r_\tau|. 
\end{equation}
Then, according to Lemma 1 in \citep{cesa2004generalization}, applying Hoeffding-Azuma inequality to the bounded variables $V_1, \dots, V_{\mu_t^u}$, we have
\begin{equation}
\frac{1}{\mu_t^u} \sum_{\tau}^{\mu_t^u}  \bbe_{(\bx, r) \sim \cald_u}[ |f(\bx; \widehat{\theta}^u_{\tau-1}) - r| ] \leq  \frac{1}{\mu_t^u} \sum_{\tau}^{\mu_t^u} | f(\bx_{\tau}; \widehat{\theta}^u_{\tau-1}) - r_\tau| + (1 + \xi_1)\sqrt{\frac{2 \log(1/\delta)}{\mu^u_t}}
\end{equation}
Because $\theta^u_{t-1}$ is uniformly drawn from $\{ \widehat{\theta}_\tau^u \}_{\tau = 0}^{\mu^u_{t-1}}$, we have
\begin{equation}\label{eq:singleg1}
\begin{aligned}
\underset{\theta^u_{t-1} \sim  \{ \widehat{\theta}_\tau^u \}_{\tau = 0}^{\mu^u_{t-1}} }{ \underset{(\bx, r) \sim \cald_u }{\bbe}}   [ |f(\bx; \theta^u_{t-1}) - r| ] & = \frac{1}{\mu_t^u} \sum_{\tau=1}^{\mu_t^u}  \bbe_{(\bx, r) \sim \cald_u}[ |f(\bx; \widehat{\theta}^u_{\tau-1}) - r| ]  \\
& \leq \underbrace{\frac{1}{\mu_t^u} \sum_{\tau=1}^{\mu_t^u} | f(\bx_{\tau}; \widehat{\theta}^u_{\tau-1}) - r_\tau|}_{I_1} + (1 + \xi_1)\sqrt{\frac{2 \log(1/\delta)}{\mu^u_t}}
\end{aligned}
\end{equation}
For $I_1$, we have
\begin{equation}\label{eq:singleg2}
\begin{aligned}
\frac{1}{\mu_t^u} \sum_{\tau=1}^{\mu_t^u} | f(\bx_{\tau}; \widehat{\theta}^u_{\tau-1}) - r_\tau|  & \underbrace{\leq}_{I_2} \frac{1}{\mu_t^u} \sum_{\tau=1}^{\mu_t^u} | f(\bx_{\tau}; \widehat{\theta}^u_{\mu^u_t}) - r_\tau| + \frac{3L}{\sqrt{2\mu^u_{t}}} \\
& \leq \frac{1}{\mu_t^u} \sqrt{\mu_t^u} \sqrt{\sum_{\tau=1}^{\mu_t^u} ( f(\bx_{\tau}; \widehat{\theta}^u_{\mu^u_t}) - r_\tau)^2}  +\frac{3L}{2\sqrt{\mu^u_{t}}}\\
& \underbrace{\leq}_{I_3} \sqrt{\frac{2\epsilon_1}{\mu_t^u}} + \frac{3L}{\sqrt{\mu^u_{t}}}.
\end{aligned}
\end{equation}
where $I_2$ is because of Lemma \ref{lemma:caogeneli} and $I_3$ is the direct application of Lemma \ref{lemma:theorem1allenzhu} (2).

Combing Eq.(\ref{eq:singleg1}) and Eq.(\ref{eq:singleg2}), we have
\begin{equation}
\underset{\theta^u_{t-1} \sim \{ \widehat{\theta}_\tau^u \}_{\tau = 0}^{\mu^u_{t-1}} }{ \underset{(\bx, r) \sim \cald_u }{\bbe}}   [ |f(\bx; \theta^u_{t-1}) - r| ] \leq \sqrt{\frac{2\epsilon_1}{\mu_t^u}} + \frac{3L}{\sqrt{2\mu^u_{t}}} + (1 + \xi_1)\sqrt{\frac{2 \log(1/\delta)}{\mu^u_t}}.
\end{equation}
Then, applying the union bound, for any $i \in [k], \tau \in [\mu_t^u ]$, the proof is completed. 
\end{proof}

\begin{lemma} \label{lemma:xi1}
 Suppose $m, \eta_1, \eta_1$ satisfy the conditions in Eq. (\ref{eq:conditions1}).  With probability at least $1 - \delta$, for any $\bx$ with $\|\bx\|_2 = 1$ and $t \in [T], u \in N$, it holds that
\[
| f(\bx; \widehat{\theta}_t^u)| \leq 2+  \mathcal{O} \left( \frac{t^4n L \log m}{ \rho \sqrt{m} } \right) + \mathcal{O} \left(  \frac{t^5 n L^2  \log^{11/6} m}{ \rho  m^{1/6}}\right) = \xi_1.
\]
\end{lemma}

\begin{proof} 
Let $\theta_0$ be randomly initialized. Then applying  Lemma \ref{lemma:gu},  for any $\bx \sim \mathcal{D}, \|\bx \|_2 = 1$ and $ \|\widehat{\theta}_t^u - \theta_0\| \leq w$,  we have 
\begin{equation}
\begin{aligned}
|f(\bx; \widehat{\theta}_t^u) | &\leq  \underbrace{| f(\bx; \theta_0)|}_{I_1} +  |\langle  \triangledown_{\theta_0}f(\bx_i; \theta_0), \widehat{\theta}_t^u - \theta_0   \rangle  | +   \mathcal{O} ( L^2 \sqrt{m \log(m)}) \|\widehat{\theta}_t^u - \theta_0 \|_2 w^{1/3} \\
&  \leq   \underbrace{2 \|\bx\|_2}_{I_1}  +  \underbrace{  \|  \triangledown_{\theta_0}f(\bx_i; \theta_0)  \|_2 \| \widehat{\theta}_t^u - \theta_0  \|_2}_{I_2}  + \mathcal{O} ( L^2 \sqrt{m \log(m)}) \underbrace{\|\widehat{\theta}_t^u - \theta_0 \|_2 w^{1/3}}_{I_3} \\
& \leq 2 + \underbrace{  \mathcal{O} (L) \cdot  \mathcal{O}\left( \frac{ t^3 }{\rho \sqrt{m}} \log m  \right)}_{I_2} + \underbrace{   \mathcal{O} \left( L^2 \sqrt{m \log(m)} \right) \cdot  \mathcal{O}\left( \frac{ t^3 }{\rho \sqrt{m}} \log m  \right)^{4/3} }_{I_3}  \\
&  =  2 +  \mathcal{O} \left( \frac{t^3 L \log m}{ \rho \sqrt{m} } \right) + \mathcal{O} \left(  \frac{t^4 L^2  \log^{11/6} m}{ \rho  m^{1/6}}\right)
\end{aligned}
\end{equation}
where $I_1$ is an application of   Lemma 7.3 in \citep{allen2019convergence}, $I_2$ is by Lemma \ref{lemma:initilizebound} (1) and  Lemma \ref{lemma:theorem1allenzhu} (4), and $I_3$ is due to Lemma \ref{lemma:theorem1allenzhu} (4).
\end{proof}

\begin{lemma} \label{lemma:caogeneli}
For any $\delta \in (0, 1)$, suppose
\[
m > \tilde{\mathcal{O}}\left(  \text{poly} (T, n, \delta^{-1}, L) \cdot  \log (1/\delta) \cdot e^{\sqrt{ \log 1/\delta}} \right), \ \ \nu = \boldsymbol{\Theta}((\mu^u_t)^6/\delta^2).  
\]
Then, with probability at least $1 - \delta$, set $\eta_1 =  \Theta( \frac{ \nu}{\sqrt{2} \mu^u_t m})$ for algorithm \ref{alg:main}-\ref{alg:user},  such that
\[
\begin{aligned}
 \sum_{\tau=1}^{\mu_t^u} | f(\bx_{\tau}; \widehat{\theta}^u_{\tau-1}) - r_\tau|  & \leq \sum_{\tau =1 }^{\mu_t^u} | f(\bx_{\tau}; \widehat{\theta}^u_{t}) - r_\tau| + \frac{3L\sqrt{\mu^u_{t}} }{\sqrt{2}} \\
\end{aligned}
\]
\end{lemma}
\begin{proof}
This is a direct application of Lemma 4.3 in \citep{cao2019generalization} by setting the loss as $L_\tau(\mathbf{W^{(\tau)}}) = | f(\bx_{\tau}; \widehat{\theta}^u_{\tau-1}) - r_\tau|  $, $R = \frac{(\mu_t^u)^3}{\delta} \log m,  \epsilon = \frac{LR}{ \sqrt{2\nu \mu_t^u }}$,  and $ \nu = R^2$, accoding to $  \| \widehat{\theta}^u_t - \widehat{\theta}^u_0\|_2 \leq \mathcal{O}\left( \frac{ (\mu^u_t) ^3 }{\delta \sqrt{m}} \log m  \right)$ (Lemma \ref{lemma:theorem1allenzhu} (3)).

\end{proof}

\begin{lemma} [Theorem 1 in \citep{allen2019convergence}] \label{lemma:theorem1allenzhu}
For any $ 0 < \epsilon_1 \leq 1$, $ 0< \rho \leq \calo(1/L)$. Given a user $u$, the collected data $\{\bx_{\tau}^u, r_{\tau}^u\}_{\tau =1}^{\mu^u_t}$, suppose $m, \eta_1, J_1$ satisfy the conditions in Eq.(\ref{eq:conditions1}).
Define $\call \left(\theta^u  \right) =  \frac{1}{2} \sum_{(\bx, r) \in  \calt^u_{t}  }  ( f(\bx; \theta^u) - r)^2$.
Then with probability at least $1 - \delta$, these hold that:
\begin{enumerate}
    \item For any $j \in [J]$, $\call(\theta^u_{(j)}) \leq (1 - \Omega(\frac{\eta_1 \rho m}{ {\mu^u_t}^2})) \call(\theta^u_{(j-1)}) $
    \item $\call(\widehat{\theta}^u_{\mu_t^u}) \leq \epsilon_1$ in $J_1 = \frac{ \text{poly}(\mu^u_t, L)}{\rho^2} \log(1/\epsilon_1) $ rounds.
    \item $\call(\theta^u_{0})  \leq  \calo( \mu_t^u \log^2m)$.
    \item For any $j \in [J]$, $\| \theta^u_{(j)}  -  \theta^u_{(0)}   \|_2 \leq \calo \left( \frac{ (\mu_t^u)^3}{ \rho \sqrt{m}} \log m \right)$.
\end{enumerate}
\end{lemma}

\begin{lemma} [Lemma 4.1, \citep{cao2019generalization}]  \label{lemma:gu} 
Suppose $ \mathcal{O}( m^{-3/2} L^{-3/2} [ \log (TnL^2/\delta) ]^{3/2}  )   \leq w \leq  \mathcal{O} (L^{-6}[\log m]^{-3/2} )$. Then,
with probability at least $1- \delta$ over randomness of $\theta_0$, for any $ t \in [T], \|\bx \|_2 =1 $, and $\theta, \theta'$
satisfying $\| \theta - \theta_0 \| \leq w$ and $\| \theta' - \theta_0 \| \leq w$
, it holds uniformly that
 \[
| f(\bx; \theta) - f(\bx; \theta') -  \langle  \triangledown_{\theta'}f(\bx; \theta'), \theta - \theta'    \rangle    | \leq \mathcal{O} (w^{1/3}L^2 \sqrt{m \log(m)}) \|\theta - \theta'\|_2.
\]
\end{lemma}

\begin{lemma} \label{lemma:initilizebound}
For any $\delta \in (0,1)$, suppose $m, \eta_1, J_1$ satisfy the conditions in Eq.(\ref{eq:conditions1}) and $\theta_0$ are randomly initialized. Then, with probability at least $1-\delta$, for any $\|\bx \|_2 = 1$, these hold that
\begin{enumerate}
    \item  $\|   \triangledown_{\theta_0 }f(\bx; \theta_0)  \|_2 \leq \mathcal{O} (L), $  \    \   \
    \item $  |  f(\bx; \theta_0)  | \leq  2 $.
\end{enumerate}
\end{lemma}

\begin{proof}
For (2), based on Lemma 7.1 in \cite{allen2019convergence}, we have $ |  f(\bx; \theta_0)  | \leq 2$. Denote by $D$ the ReLU function.  For any $l \in [L]$, 
\[
\|\triangledown_{W_l}f(\bx; \theta_0)\|_F \leq   \|\bw_L D \bw_{L-1} \cdots D \bw_{l+1}\|_F \cdot \| D \bw_{l+1} \cdots \bx \|_F \leq  \mathcal{O}(\sqrt{L}) 
\] 
where the inequality is according to Lemma 7.2 in \cite{allen2019convergence}. Therefore, we have $ \|   \triangledown_{\theta_0 }f(\bx; \theta_0)  \|_2 \leq \mathcal{O} (L) $. 
\end{proof}

\end{document}